\documentclass[10pt,twocolumn,letterpaper]{article}
\usepackage[pagenumbers]{cvpr}

\usepackage{booktabs}
\usepackage{multirow}
\usepackage{placeins}
\definecolor{cvprblue}{rgb}{0.21,0.49,0.74}
\usepackage[breaklinks,colorlinks,allcolors=cvprblue]{hyperref}

\title{Queryable Gaussian View Index:\\Retrieval, Localization, and Agentic View Search}
\author{%
  Tianyu Xiong$^{1}$\qquad Rui Li$^{2}$\qquad Jiaqi Yang$^{1}$\\[0.6em]
  \normalsize $^{1}$School of Computer Science, Northwestern Polytechnical University, Xi'an, China\\[0.35em]
  \normalsize $^{2}$CEMSE, King Abdullah University of Science and Technology, Thuwal, Saudi Arabia
}
\date{}
\hypersetup{
  pdftitle={Queryable Gaussian View Index: Retrieval, Localization, and Agentic View Search},
  pdfauthor={Tianyu Xiong, Rui Li, Jiaqi Yang},
  pdfsubject={Text retrieval, camera localization, and agentic view search in Gaussian scene assets},
  pdfkeywords={3D Gaussian splatting, text retrieval, visual localization, agentic view search}
}
\makeatletter
\apptocmd{\@maketitle}{\par\noindent
\begin{minipage}{\textwidth}
  \centering
  \includegraphics[width=\linewidth]{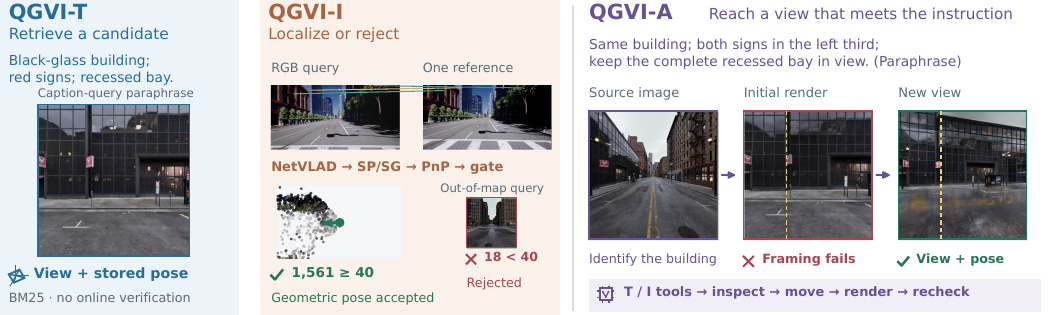}
  \captionsetup{type=figure,hypcap=false}
  \caption{\textbf{Queryable Gaussian View Index (QGVI).}
  QGVI-T returns caption-ranked views and stored poses (left).
  QGVI-I gates geometric LoGS poses (middle); the plot
  shows the accepted pose on SfM points. QGVI-A uses posed candidates and
  camera exploration to satisfy an image-relative instruction (right).
  Selected cached synthetic-scene observations omit intermediate proposals;
  descriptions are paraphrased and semantic checks are model assessments.}
  \label{fig:teaser}
\end{minipage}
\par
}{}{%
  \PackageError{qgvi}{Could not attach the teaser to the title block}{}
}
\makeatother

\begin{document}
\maketitle
\begin{abstract}
Given an existing 3D Gaussian scene and calibrated references, we seek camera
poses and views that answer text, RGB, or image-and-instruction queries.
RGB localization recovers a source camera; a target-view instruction can
require a different camera while preserving the referenced scene content.
Queryable Gaussian View Index (QGVI) supports these requests through a
semantic view index that binds structured captions and observations to
calibrated camera poses. QGVI-T retrieves indexed candidates with BM25.
QGVI-A uses this interface and geometric source localization (QGVI-I) as
tools, sharing candidate poses, observations, and query conditions marked
satisfied, violated, or unknown. Unmet conditions guide camera proposals
and rendered checks. Supported answers exit early; unresolved queries
return insufficient evidence within a shared search budget.
We construct a nine-scene benchmark, including city-scale environments,
covering semantic view retrieval, RGB localization with out-of-map rejection,
and image-conditioned target-view search.
Under GPT assessment on the development cohort, QGVI-A returns confirmed
correct views for
90.4\% of target-view queries, compared with 26.8\% for one-shot
Retrieve + Select. This comparison evaluates complete policies with
different information and call budgets; QGVI-A achieves higher coverage
at lower answer precision. The benchmark separates semantic satisfaction,
source-camera accuracy, and acceptance decisions.

\end{abstract}

\section{Introduction}
\label{sec:introduction}

Given a Gaussian reconstruction and calibrated references, we seek
scene-frame camera poses and views answering a user's query. Text requests
views satisfying a description; RGB requests its source camera. An image
with an instruction can request a different camera: in Fig.~\ref{fig:teaser},
the building must remain the same while its signs appear in the left third
and the recessed bay is fully visible. Source localization identifies the
camera, but does not establish that the target conditions hold.

Composed image retrieval selects a dataset image using a reference image
and modifying text~\cite{liu2021cirr}. CityLoC estimates text-conditioned pose
distributions in Gaussian scenes~\cite{ma2025cityloc}, accommodating multiple
plausible answers. Our queries return satisfying camera views within the
supplied asset, separating semantic satisfaction from source-camera accuracy.
The Gaussian reconstruction provides the scene and a renderer for inspecting
proposed cameras~\cite{kerbl2023gaussians}.

Queryable Gaussian View Index (QGVI) starts with a semantic view index.
Each entry binds an original or rendered observation to a calibrated
camera and structured caption describing objects, attributes, relations,
and visible text. QGVI-T uses BM25 to retrieve candidates after removing
duplicate images and poses. This caption-based image retrieval interface
provides cameras for further search. Captioning is offline;
retrieval requires no online VLM call or scene-specific pose network.
Grid renders are optional; their retrieval benefit is not established.

QGVI-A uses QGVI-T and QGVI-I as tools. QGVI-I supplies LoGS geometric
poses~\cite{cheng2025logs}, accepted according to distinct supporting query
keypoints without changing accepted estimates. The agent maintains candidate
poses, observations, and task-supplied conditions marked satisfied, violated,
or unknown. Unmet conditions guide further tool calls and camera proposals;
renders supply the next observation. Supported answers exit early;
unresolved queries return insufficient evidence when the budget is exhausted.
The tools share coordinates and observations; a query need not call both.

Our benchmark covers all three query regimes over nine existing scenes,
including city-scale environments: 180 text queries, 1,839 in-map and 1,839
out-of-map RGB queries, and 250 image-and-instruction queries from 125 source
images. The query cohorts and protocols measure semantic satisfaction,
pose accuracy, correct/wrong returns, and coverage over shared assets.
These development evaluations use model semantic assessments; target-view
controls compare whole policies with different information and call budgets.

Our contributions are:
\begin{itemize}
  \item A semantic view index linking structured captions and observations
  to calibrated camera poses for retrieval and subsequent scene queries.
  \item Agentic localization and view search using QGVI-T and QGVI-I as
  tools, with shared evidence, condition-guided camera proposals, early
  exit, and budgeted abstention.
  \item A nine-scene multimodal benchmark covering text-view retrieval,
  RGB source localization, and image-conditioned target-view search,
  with task-specific correctness and common answer accounting.
\end{itemize}

\section{Related Work}
\label{sec:related_work}

\paragraph{Text-conditioned localization.}
Text2Pos and Text2Loc localize descriptions within point-cloud maps
through coarse-to-fine inference~\cite{kolmet2022text2pos,xia2024text2loc};
PMSH models uncertainty when descriptions only partially match
submaps~\cite{feng2025pmsh}. CityLoC~\cite{ma2025cityloc} instead learns a
text-conditioned camera-pose distribution using diffusion and
rendered candidates for refinement, accommodating multiple plausible poses.
Our semantic query task asks for a bounded set of satisfying views; source-pose
error alone cannot credit every valid alternative. This task distinction
does not establish that discrete retrieval universally outperforms continuous
estimation.

\paragraph{Image retrieval and semantic scene representations.}
CLIP~\cite{radford2021clip} retrieves through aligned image--text embeddings.
QGVI-T is caption-based image retrieval using existing
BM25~\cite{robertson2009bm25} over structured descriptions.
Its semantic view index binds each observation and caption to a scene-frame
camera pose, providing candidates for rendering and subsequent agentic search.
Historical fusion and matched controls also evaluate MiniLM-based
sentence embeddings~\cite{wang2020minilm,reimers2019sentencebert}.
Description satisfaction and source-camera recovery are evaluated separately.

OpenScene aligns point features with image and text
embeddings~\cite{peng2023openscene}. LERF, LangSplat, and Feature 3DGS
extend language or semantic features to renderable scene
representations~\cite{kerr2023lerf,qin2024langsplat,zhou2024feature3dgs}.
QGVI-T indexes view-level descriptions without learning a 3D language field.
Composed image retrieval, including CIRR~\cite{liu2021cirr}, combines a reference
image and modifying text to retrieve another dataset image. Our target-view
queries retain this combined input but require a camera pose and rendered
observation within the supplied scene, subject to source-relative predicates.

\paragraph{Geometric localization.}
Structure-based localization uses retrieval and matching to establish
image-to-map correspondences.
Active Search prioritizes direct 2D--3D matching~\cite{sattler2017activesearch};
hierarchical localization retrieves candidate places before local
matching~\cite{sarlin2019hloc}. InLoc further combines dense matching with
synthesized-view verification~\cite{taira2018inloc}.

NetVLAD, MixVPR, and SALAD retrieve
reference images~\cite{arandjelovic2016netvlad,alibey2023mixvpr,izquierdo2024salad}.
Local features include SIFT, RootSIFT
normalization~\cite{lowe2004sift,arandjelovic2012rootsift}, and learned
SuperPoint and
D2-Net~\cite{detone2018superpoint,dusmanu2019d2net}.
SuperGlue and LightGlue assign sparse feature
matches~\cite{sarlin2020superglue,lindenberger2023lightglue}; LoFTR
establishes detector-free correspondences~\cite{sun2021loftr}.
COLMAP supplies sparse landmarks~\cite{schonberger2016sfm};
PnP solvers such as EPnP~\cite{lepetit2009epnp} and robust consensus
estimation~\cite{fischler1981ransac} recover poses from noisy matches.
Scene-coordinate regression learns image-to-scene associations directly:
DSAC makes robust estimation differentiable~\cite{brachmann2017dsac},
and ACE accelerates scene-specific training~\cite{brachmann2023ace}.
QGVI-I adopts NetVLAD, SuperPoint, SuperGlue, and SfM landmarks through
LoGS~\cite{cheng2025logs}. Its added acceptance rule counts distinct inlier
query keypoints at LoGS's geometric pose, avoiding repeated support across
references, and leaves that pose fixed.

\paragraph{Rendering and pose refinement.}
NeRF represents continuous radiance fields~\cite{mildenhall2020nerf};
iNeRF refines an initial pose using rendered-image
residuals~\cite{yenchen2021inerf}. Feature-based alternatives include
PixLoc's learned multiscale alignment~\cite{sarlin2021pixloc} and MCLoc's
pretrained features with particle filtering~\cite{trivigno2024mcloc}.
3D Gaussian Splatting supports differentiable novel-view
rendering~\cite{kerbl2023gaussians}. GSLoc differentiates image-alignment
error with coarse-to-fine blurring~\cite{botashev2024gsloc}; LoGS adds
affine brightness correction and informative-pixel masks to photometric
refinement~\cite{cheng2025logs}. GSplatLoc grounds keypoint descriptors
in Gaussian maps~\cite{sidorov2025gsplatloc}, and SplatHLoc combines
rendered and image-derived features at different matching
stages~\cite{tao2026splathloc}. 6DGS estimates pose from image--ray
bindings without requiring an initial pose or iterative pose
optimization~\cite{bortolon2024sixdgs}.
Rendering optionally supplies QGVI-T's grid views. We evaluate the released
LoGS photometric optimizer as an ablation of the geometric default.
QGVI-A uses rendered feedback to inspect proposed camera
poses against supplied predicates, including source-relative constraints.

\paragraph{Selective prediction.}
Selective classification withholds predictions, exposing a
risk--coverage trade-off~\cite{geifman2017selective}. For
localization, accepting an out-of-map image, rejecting an in-map image,
and accepting an inaccurate in-map pose are distinct outcomes. QGVI-I
selects its threshold subject to retaining validation-correct poses and
holds the underlying estimates fixed in the paired gate ablation.
Historical QGVI-T verification withholds unsupported candidates. QGVI-A can
search further through camera proposals, terminating early on supported
answers or abstaining within a fixed budget. These decisions are evaluated
separately from view relevance and pose accuracy.

\section{Method}
\label{sec:method}

\begin{figure*}[t]
  \centering
  \includegraphics[width=\textwidth]{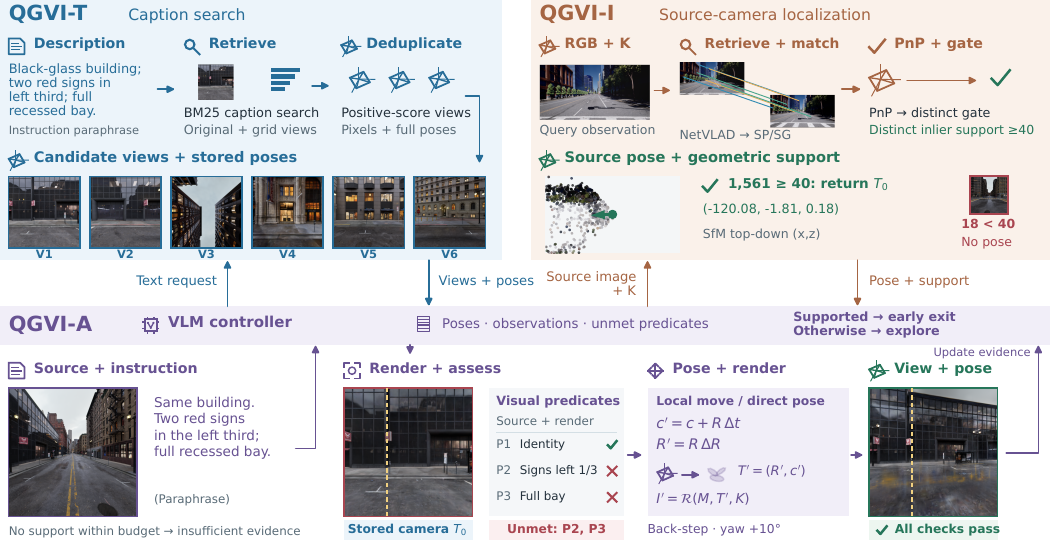}
  \caption{\textbf{QGVI-A coordinates caption search and source localization.}
  Call/return links expose tool interfaces; Sec.~\ref{sec:experiments} specifies
  the evaluated variants. Cached BM25 candidates from the evaluated
  original-plus-grid bank seed the pictured episode.
  The agent checks supplied predicates, updates the camera $T=(R,c)$, and
  renders $\mathcal{R}(M,T,K)$ with fixed intrinsics $K$. The equations show
  local camera increments; direct pose proposals are also supported.
  The RGB example shows a geometric LoGS pose on SfM points.
  Selected observations omit intermediate proposals; guides mark the left third.
  Checks are model assessments, coordinates use native units, and images depict
  synthetic scenes.}
  \label{fig:overview}
\end{figure*}

\subsection{Task: Querying a Gaussian Scene Asset}
The input scene $M$ comprises an existing Gaussian reconstruction and its
associated calibrated reference data. A query supplies text, an RGB image
with intrinsics, or a source image with a target-view instruction and
required predicates. Outputs are scene-frame camera poses and corresponding
views, or no answer. For query $q$
and task type $\tau\in\{\mathrm{text},\mathrm{source},\mathrm{target}\}$,
define
\begin{equation}
 \mathcal A_\tau(q,M)=\{T\in SE(3):C_\tau(q,T;M)=1\}.
 \label{eq:scene_grounding}
\end{equation}
Here $T$ is a scene-frame camera pose. For text, $C_{\mathrm{text}}$ requires
a usable view satisfying the description; several distant poses may qualify.
For RGB source localization, $C_{\mathrm{source}}$ requires an in-map query
and agreement with its source pose within the evaluation thresholds.
Out-of-map RGB queries have an empty valid-answer set. For an image with a
target-view instruction, $C_{\mathrm{target}}$ requires a usable rendered
view satisfying every supplied predicate and its relation to the source;
recovering the source camera alone is insufficient. These sets define
evaluation targets and are not computed exhaustively. QGVI-T retrieves stored
poses, QGVI-I estimates a source pose, and QGVI-A coordinates their interfaces
with camera exploration. Each may return no answer.

The routes use different reference assets (Fig.~\ref{fig:overview}).
QGVI-T indexes original images and optional renders from frozen Horizon-GS
reconstructions~\cite{kerbl2023gaussians,jiang2025horizongs}, with captions
and camera-to-world poses. QGVI-I uses original training
RGB images and sparse landmarks triangulated from feature matches with
fixed training cameras~\cite{schonberger2016sfm}, using LoGS geometric
initialization~\cite{cheng2025logs} and a calibrated acceptance rule.
Supplementary
Sec.~\ref{sec:implementation_details} gives implementation settings.

\subsection{QGVI-T: Language-Indexed View Retrieval}
\label{sec:text_method}
\paragraph{Semantic view index.}
Each record binds an observation $I_i$, its camera-to-world pose $T_i$,
calibration $K_i$, and structured caption $d_i$. The index starts from
calibrated reference images. Optional augmentation
adds renders from a deterministic lattice near the training cameras,
with sampled orientations and training-camera intrinsics.
Within each scene and camera domain, views sharing a center share a place
identifier. Augmentation adds virtual observations with explicit camera
poses; dense pre-rendering is not required by the interface.

\paragraph{Caption retrieval.}
DeepSeek describes each reference offline using structured fields
for objects, attributes, counts, relations, and visible text. The prompt
prohibits inferring hidden objects or camera metadata; the uncertainty field
is excluded from the index. Reference images and poses remain fixed.

The current caption-search interface uses whole-caption
BM25~\cite{robertson2009bm25}, without embeddings. It ranks positive-score
views and removes duplicate pixels or camera poses, returning up to six
candidates with their stored cameras for inspection or further view proposals.
Ranking does not verify semantics or
establish that a matching view is absent.

\paragraph{Historical fusion variant.}
Earlier experiments combine MiniLM~\cite{wang2020minilm,reimers2019sentencebert}
and BM25 rankings by reciprocal-rank fusion~\cite{cormack2009rrf} and return
up to five distinct places. These results retain the fusion label.
Supplementary Secs.~\ref{sec:implementation_details} and
\ref{sec:supp_text_verification} detail its shortlist and optional
visual-verification ablation.

\subsection{QGVI-I: Selective Geometric Localization}
\label{sec:image_method}
\paragraph{Geometric initialization.}
For image $I$ with intrinsics $K$, the LoGS initializer retrieves ten training
images with NetVLAD~\cite{arandjelovic2016netvlad} and matches SuperPoint
features~\cite{detone2018superpoint} using
SuperGlue~\cite{sarlin2020superglue}. Reference feature indices connect
matches to triangulated landmarks. Correspondences from all references
form one PnP problem without covisibility clustering.
PnP-RANSAC~\cite{fischler1981ransac} uses a 12-pixel reprojection threshold
and is followed by geometric pose refinement, yielding $T_0$.
Our acceptance rule operates on this geometric estimate without
photometric optimization.

\paragraph{Distinct geometric support.}
Matches from several references can repeat the same query feature.
At the geometric camera-to-world pose $T_0$, we therefore count each query
keypoint once, even when it matches several references or competing landmarks.
For keypoint $u_j$, let $\mathcal{P}_j^+(T_0)$ contain its matched landmarks
with positive depth. Selecting the smallest reprojection residual gives
\begin{equation}
 n_{\mathrm{u}}(T_0)=
 \sum_j\mathbf{1}\!\left[
 \min_{X\in\mathcal{P}_j^+(T_0)}
 \|\pi_K(T_0^{-1}X)-u_j\|_2\leq12
 \right].
 \label{eq:unique_support}
\end{equation}
The minimum of an empty set is infinity, and $\pi_K$ is calibrated
projection. We reproject the cached geometric correspondences at the saved
geometric pose. Support counts distinct query keypoints, not correspondence rows.

\paragraph{Acceptance calibration.}
We calibrate on 490 Block A and SF validation queries, equally split between
in-map and out-of-map examples. A fixed count grid selects the threshold
while retaining every correct geometric validation pose separately at
1\,m/2$^\circ$ and 5\,m/5$^\circ$. Selection prioritizes out-of-map rejection,
then in-map retention, then the smaller threshold. The selected cutoff is
40 distinct inliers. Let $e_{\mathrm{PnP}}(I)$ indicate successful PnP
estimation. Frozen across nine test scenes, the rule is
\begin{equation}
 \begin{aligned}
 a(I)&=e_{\mathrm{PnP}}(I)\mathbf{1}[n_{\mathrm{u}}(T_0)\geq40],\\
 \widehat T(I)&=
 \begin{cases}T_0,&a(I)=1,\\ \varnothing,&a(I)=0.\end{cases}
 \end{aligned}
 \label{eq:image_acceptance}
\end{equation}
The gate preserves $T_0$. The optional ``QGVI-I + photo.'' ablation runs
up to 20 native LoGS photometric steps with brightness correction and
informative-pixel masks, then gates the refined pose $T_\star$ under native
LoGS eligibility. Its independently calibrated distinct-count cutoff is
also 40. Neither gate guarantees correct-pose retention on unseen queries.

\subsection{QGVI-A: Adaptive Scene Querying}
\label{sec:agent_method}
\paragraph{Shared evidence and constraints.}
QGVI-A calls QGVI-T's caption-search and QGVI-I's source-localization
interfaces (Fig.~\ref{fig:overview}) using shared candidates, camera poses,
observations, geometric support, and unmet task-supplied predicates.
RGB calls use the source image; text calls retrieve candidate cameras.
The controller inspects target renders and records each predicate as satisfied,
violated, or unknown, citing observations. Acceptance requires every predicate
and the source relation to be supported by these model assessments.

\paragraph{Exploration and stopping.}
An ordinary retrieval/localization prefix first attempts a direct answer:
a source pose must pass the geometric gate; a target view must pass the
rendered-predicate check. Otherwise, its evidence enters adaptive search.
For camera-to-world pose $T=(R,c)$, local actions update
$c'=c+R\Delta t$ and $R'=R\Delta R$; the controller may also propose a
bounded valid $SE(3)$ pose. Rendering $I'=\mathcal R(M,T',K)$ with fixed
intrinsics supplies the next observation. Inspection and camera updates
continue within shared decision, render, and time budgets. At termination,
the host returns a saved supported answer or insufficient evidence.
Sec.~\ref{sec:experiments} and Supplementary
Sec.~\ref{sec:supp_selective_answers} specify evaluated tool variants and budgets.

\section{Experiments}
\label{sec:experiments}

The development benchmark contributes cohorts and protocols for text retrieval,
RGB localization, target-view search, rejection, and latency over nine existing
scenes, including city-scale environments.

\begin{table*}[tp]
 \centering
 \caption{\textbf{Shared answer accounting and net query utility.}
 First-answer results pooled over nine scenes. $N$ counts query--seed observations
 for CityLoC and queries otherwise. Net utility is $100(S-W)/N$.
 Semantic correctness requires a usable view satisfying the description or
 all target predicates. RGB correctness requires an in-map pose within
 1\,m/2$^\circ$; five scenes have DA3-estimated scales.
 Ranges propagate uncertain GPT judgments, not sampling uncertainty.
 Panel (a) uses the common GPT-low relevance/validity rubric.
 Panels use different cohorts or judging protocols and must not be ranked
 against each other. Semantic panels bold best conservative endpoints; RGB
 bold marks best values, including ties. Coverage is descriptive.}
 \label{tab:main}
 \label{tab:selective_answers}
 \small
 \setlength{\tabcolsep}{4pt}
 \begin{tabular}{@{}lrrrrrr@{}}
 \toprule
 Method & $N$ & \shortstack{Answered\\$C$ (\%)} &
 \shortstack{Answer precision\\$P$ (\%) $\uparrow$} &
 \shortstack{Correct returns\\$Y$ (\%) $\uparrow$} &
 \shortstack{Wrong returns\\$E$ (\%) $\downarrow$} &
 \shortstack{Net utility\\$100U_1$ $\uparrow$} \\
 \midrule
 \multicolumn{7}{l}{\textit{(a) Text: common GPT-low relevance and validity judgments}} \\
CityLoC w/o refinement & 540 & 100.00 & 26.67--66.48 & 26.67--66.48 & 33.52--73.33 & -46.67--32.96 \\
CityLoC with refinement & 540 & 80.19 & 36.26--61.66 & 29.07--49.44 & 30.74--51.11 & -22.04--18.70 \\
QGVI-T & 180 & 100.00 & 60.00--70.56 & 60.00--70.56 & 29.44--40.00 & 20.00--41.11 \\
QGVI-A & 180 & 72.78 & \textbf{90.84--93.89} & \textbf{66.11--68.33} & \textbf{4.44--6.67} & \textbf{59.44--63.89} \\
\midrule
\multicolumn{7}{l}{\textit{(b) RGB: geometric correctness, 50\% out-of-map}} \\
LoGS, native & 3,678 & 100.00 & 48.26 & \textbf{48.26} & 51.74 & -3.48 \\
SplatHLoc, calibrated count & 3,678 & 99.54 & 30.73 & 30.59 & 68.95 & -38.36 \\
QGVI-I & 3,678 & 49.78 & 96.94 & \textbf{48.26} & 1.52 & 46.74 \\
QGVI-I + photo. & 3,678 & 49.70 & \textbf{97.10} & \textbf{48.26} & \textbf{1.44} & \textbf{46.82} \\
QGVI-A & 3,678 & 49.78 & 96.83 & 48.21 & 1.58 & 46.63 \\
\midrule
\multicolumn{7}{l}{\textit{(c) Image-and-instruction target views: GPT high-effort judgments}} \\
Retrieve + Select & 250 & 27.20 & \textbf{98.53} & 26.80 & \textbf{0.40} & 26.40 \\
Fixed Search + Verify & 250 & 20.00 & 88.00--90.00 & 17.60--18.00 & 2.00--2.40 & 15.20--16.00 \\
QGVI-A & 250 & 97.60 & 92.62--93.03 & \textbf{90.40--90.80} & 6.80--7.20 & \textbf{83.20--84.00} \\

 \bottomrule
 \end{tabular}
\end{table*}

\subsection{Experimental Setup}
\label{sec:exp_setup}

\paragraph{Datasets and reference views.}
The nine-scene benchmark combines Block Small and Block A from
MatrixCity~\cite{li2023matrixcity}, NYC and SF from
UC-GS~\cite{zhang2024ucgs}, and Citysample, Colosseum, Elvenruin, Park,
and Road from the Horizon-GS dataset~\cite{jiang2025horizongs}.
Seven scenes are synthetic; Park and Road contain real captures.
We use frozen Horizon-GS reconstructions across all nine scenes to
generate reference views and novel-view queries.
The BM25 bank contains 18,570 original-image entries and 30,522 grid renders
(49,092 entries); rendered-bank controls use the latter. The source split has
no exact file overlap, but some evaluation observations contributed to reconstruction.

\paragraph{Query cohorts.}
The main text cohort contains 180 DeepSeek-authored descriptions,
20 per scene: 74 from original test images and 106 from novel
Gaussian renders; six are unsupported by their source images. Earlier
176-query and 135-query cohorts remain separate.

RGB testing uses 1,839 in-map and 1,839 out-of-map queries. Positives
comprise 1,029 held-out source images, including synthetic imagery, and
810 novel Gaussian renders. Negatives comprise 1,029 training images and
810 renders from disjoint scene worlds within the nine scenes. Query
composition varies by scene; Block A supplies 32.7\% of positives.
Methods share frozen queries, splits, calibration, and target scenes.
Four scene scales are source-verified
and five DA3-estimated~\cite{lin2025depthanything3}
(Supplementary Table~\ref{tab:nine_scene_detail}); calibration and interpretation are specified in
Supplementary Sec.~\ref{sec:supp_protocol}.

\paragraph{Compared methods.}
Our CityLoC~\cite{ma2025cityloc} reimplementation uses three fixed seeds
and 64 pose proposals. QGVI-T returns the first BM25 candidate from the
original-plus-grid bank; QGVI-A returns its saved final answer. Historical
fusion and verification controls are retained in the supplement.
QGVI-I uses the LoGS~\cite{cheng2025logs} geometric initializer and accepts
successful PnP poses with at least 40 distinct inlier query keypoints.
The optional ablation adds native photometric refinement (``+ photo.'').
QGVI-I shares LoGS's native assets; SplatHLoc~\cite{tao2026splathloc}
also uses its native map.
Native LoGS accepts retrieval fallbacks; SplatHLoc requires four final
inliers. We calibrate count and ratio gates for both LoGS pose variants,
and a count gate for SplatHLoc on the same 490 Block A/SF validation
queries, preserving correct poses, then freeze thresholds across nine scenes.

\paragraph{Model judgments.}
The main text comparison uses one GPT-6.1 Sol low-effort relevance/validity
rubric for all frozen first answers. Compatible historical judgments are
reused; new reviews preserve query text and image bytes. Request transports
differ, as disclosed in Supplementary Sec.~\ref{sec:supp_selective_answers}.
DeepSeek supplies captions and descriptions. Earlier Qwen/DeepSeek and
Kimi/GPT assessments remain separate (Supplementary Sec.~\ref{sec:supp_human}).
Matched retrieval controls retain their own judging protocol.

\paragraph{Agentic evaluation.}
QGVI-A uses early exit, bounded search, BM25 over the original-plus-grid
bank, and geometric QGVI-I. Alongside text/RGB, we test 250
image-and-instruction queries from 125 source images. \emph{Retrieve + Select}
makes one choice among the same six initial BM25 renders. \emph{Fixed Search + Verify} checks six
precommitted cameras around the first candidate, stopping at the first
supported view. All use GPT-6.1 Sol low and at most six render attempts,
with separate high-effort predicate assessments. Information access and
call budgets differ (Supplementary Secs.~\ref{sec:supp_selective_answers}
and~\ref{sec:target_baselines}); text retains its common low-effort rubric.

\paragraph{Common answer metrics.}
Let $A$ count first returns among $N$ queries, with $S$ correct and $W$ wrong.
Coverage is $C=A/N$, answer precision $P=S/A$,
correct-return rate $Y=S/N$, and wrong-return rate $E=W/N$.
Uncertain judgments bound $S$ and $W$ without changing $A$.
Net query utility $U_\lambda=Y-\lambda E$ assigns reward $+1$ to a
correct return, $-\lambda$ to a wrong return, and zero to no return
\cite{franc2023reject}. Table~\ref{tab:main} uses $U_1=Y-E$;
equal reward and penalty are illustrative, not calibrated deployment costs.
Correctness follows the panel-specific criteria in the caption.
Utility does not replace pose error or match acceptance policies;
Supplementary Sec.~\ref{sec:supp_selective_answers} gives cost sensitivity.

\paragraph{Text metrics.}
Top-$K$ success (Hit@$K$) counts queries with a valid image satisfying all
textual constraints among the first $K$ outputs; empty lists fail.
The 176-query cohort measures relevance only. Lower bounds count definite
successes; upper bounds also count uncertain outcomes. These are label
bounds, not confidence intervals or full-corpus recall. No-result rate
counts empty returns, excluding runtime failures and missing judgments;
it cannot certify that no matching view exists. Scene-macro scores weight
scenes equally after CityLoC seed averaging; pooled scores weight queries.

\paragraph{RGB metrics.}
An in-map query depicts the target scene; an out-of-map query depicts
another scene world. Translation error is camera-center distance;
rotation error is the relative rotation angle. Pose success divides
accepted poses within both $1$\,m/$2^\circ$ (strict) or $5$\,m/$5^\circ$
(loose) by \emph{all in-map queries}, including rejected or missing poses
as failures. Metre thresholds include five estimated scales.
With in-map indicator $y$ and acceptance indicator $a$,
\begin{equation}
  \mathrm{FAR}=\frac{\sum(1-y)a}{\sum(1-y)},\qquad
  \mathrm{FRR}=\frac{\sum y(1-a)}{\sum y}.
\end{equation}
FAR and FRR use all out-of-map and in-map queries, respectively, and are
independent of metre scale. Accepted wrong in-map poses are localization
failures. Accepted-pose precision divides correct poses by all accepted
outputs; conditional in-map precision excludes out-of-map outputs.
Precision depends on the query mixture, balanced here.

\subsection{Answer Quality and Localization Results}
\label{sec:exp_main}

\paragraph{First-answer utility.}
Table~\ref{tab:main} compares text methods under one judging rubric.
QGVI-A yields 59.44--63.89 net reward units per
100 text queries; BM25 QGVI-T yields 20.00--41.11.
On target-view queries, QGVI-A yields 83.20--84.00,
versus 26.40 for Retrieve + Select and 15.20--16.00 for Fixed Search + Verify.
Confirmed correct-return rates are 90.4\%, 26.8\%, and 17.6\%, respectively.
The one-shot control has higher precision (98.53\%) at 27.2\% coverage;
QGVI-A answers 97.6\% at 92.62--93.03\% precision. This compares complete
policies, not navigation alone. Supplementary Sec.~\ref{sec:target_baselines}
gives scene, task-family, and paired uncertainty analyses.
On RGB, QGVI-A yields 46.63, below QGVI-I's 46.74 and its photometric
ablation's 46.82. Its 1,831 returns comprise 1,773 strict successes,
46 inaccurate in-map poses, and 12 out-of-map accepts.

\paragraph{Text-view retrieval.}
BM25 QGVI-T returns 180/180 answers, with 108 confirmed correct,
53 incorrect, and 19 uncertain. QGVI-A returns 131/180,
with 119 confirmed correct, 8 incorrect, and 4 uncertain.
Supplementary Table~\ref{tab:gpt_low_text_all} includes fusion and verification
under the same rubric; Table~\ref{tab:historical_text_utility} preserves earlier
judgments. All assessments use frozen localization outputs.

\paragraph{Source poses and semantic judgments.}
On these 180 descriptions, QGVI-T fusion achieves 0.6\% Top-1 source-pose
success and 3.3\% Pose Recall@5 at 5\,m/5$^\circ$; full-bank oracle
coverage is 22.2\%. Semantic success does not establish source-camera accuracy
(Supplementary Sec.~\ref{sec:supp_additional}).

\paragraph{RGB localization and acceptance.}
QGVI-I achieves 97.3\% and 97.4\% scene-macro pose success at
1\,m/2$^\circ$ and 5\,m/5$^\circ$
(Supplementary Table~\ref{tab:nine_scene_detail}(b)). It accepts
13/1,839 out-of-map queries (0.707\% pooled; 0.764\% scene mean) and
rejects 21/1,839 in-map queries (1.142\% pooled; 0.765\% scene mean).
Accepted in-map poses include 43 strict and 40 loose localization
failures. LoGS and SplatHLoc achieve 97.3\% and 76.0\% strict success,
with 100.0\% and 99.8\% scene-mean false acceptance under native output
rules. These native rates do not measure the benefit over calibrated gates.
Every NYC and SF positive lies within 0.59\,m of a training camera,
with shared reconstruction dependencies; difficulty varies across scenes.

The QGVI-I gate preserves its 1,831 retained geometric pose matrices
(Table~\ref{tab:selective_pose}) and all 1,775 strict and 1,778 loose successes.
The photometric ablation retains these success counts and rejects three
more distractors; precision rises from 96.94\% to 97.10\%.
Under matched validation-retention objectives (Table~\ref{tab:calibrated_gates}),
raw and distinct initializer counts accept 24 and 13 distractors, respectively;
the photometric ablation's final distinct support accepts ten. All retain the same strict-success
queries, with different rejection and computational costs.

Photometric refinement changes all 3,678 pose matrices without strict or loose
success/failure transitions. In-map median
translation/rotation errors increase from 0.012\,m/0.023$^\circ$ to
0.028\,m/0.085$^\circ$. Translation and rotation errors worsen for 78.0\%
and 93.0\% of in-map queries. We therefore retain the geometric estimate
by default (Supplementary Sec.~\ref{sec:calibrated_rgb_supp}).

\begin{table}[t]
\centering
\caption{\textbf{Rejection under a shared calibration objective.}
Thresholds retain every validation-correct pose, then maximize rejection
of validation distractors. Pooled test rates (\%) use 1,839 queries per class;
pose success is at 1\,m/2$^\circ$. QGVI-I gates the geometric initializer;
``+ photo.'' adds native LoGS refinement and the final-pose gate.
SplatHLoc retains its native requirement of at least four inliers.
Bold marks the best displayed rates and ties.}
\label{tab:calibrated_gates}
\footnotesize\setlength{\tabcolsep}{3pt}
\begin{tabular}{@{}lrrr@{}}\toprule
Method / support threshold & Pose $\uparrow$ & FAR $\downarrow$ & FRR $\downarrow$ \\ \midrule
Initializer: count $\geq40$ & \textbf{96.52} & 1.305 & 1.033 \\
Initializer: ratio $\geq0.1$ & 96.36 & 79.445 & 0.761 \\
QGVI-I: distinct $\geq40$ & \textbf{96.52} & 0.707 & 1.142 \\
LoGS: pairs $\geq40$ & \textbf{96.52} & 1.251 & 1.142 \\
QGVI-I + photo.: $\geq40$ & \textbf{96.52} & \textbf{0.544} & 1.142 \\
SplatHLoc: native ($\geq4$) & 61.17 & 99.347 & \textbf{0.272} \\
\bottomrule\end{tabular}
\end{table}

\subsection{Matched Text Retrieval Controls}
\label{sec:exp_text_ablation}

Table~\ref{tab:matched_text} compares CLIP, BM25, MiniLM, and historical
fusion on the same 180 descriptions and rendered bank: 64 candidates,
a 20-view shortlist, five distinct places, and no online verification.
All 2,825 output pairs have GPT judgments. BM25's confirmed Top-1/Top-5
(62.8\%/77.2\%) exceeds fusion's (61.1\%/73.9\%); overlapping label bounds
do not establish a fusion benefit. Supplementary Sec.~\ref{sec:matched_text_supp}
gives per-scene results and partial Kimi assessments.

\begin{table}[t]
\centering
\caption{\textbf{Matched text retrieval controls.} Same 180 queries,
30,522 rendered candidates, place selection, and five-result budget;
no online verification. GPT judgments are complete. Ranges are
label-uncertainty bounds (\%), not confidence intervals. Bold marks
the largest lower bound; overlapping ranges do not establish a ranking.}
\label{tab:matched_text}
\small\setlength{\tabcolsep}{5pt}
\begin{tabular}{@{}lcc@{}}\toprule
Retrieval & Top-1 $\uparrow$ & Top-5 $\uparrow$ \\ \midrule
CLIP & 39.4--54.4 & 62.2--80.6 \\
BM25 & \textbf{62.8}--78.9 & \textbf{77.2}--92.2 \\
MiniLM & 56.7--73.9 & 70.6--90.0 \\
QGVI-T (fusion) & 61.1--77.8 & 73.9--91.7 \\
\bottomrule\end{tabular}
\end{table}

Optional verification remains a separate operating-point ablation. Under
the historical Qwen/DeepSeek rubric it raises Top-1 from 58.3\% to
66.1\% but lowers Top-5 from 78.3\% to 72.8\%, with 45/180 empty returns.
The earlier 176-query controls and query-type breakdown are retained in
Supplementary Tables~\ref{tab:text_ablation} and~\ref{tab:breakdown}.

\subsection{Contribution of Rendered Grid Views}
\label{sec:exp_view_bank}

We add 30,522 Gaussian grid renders to 18,570 original images, fixing captions,
historical fusion, queries, and the five-result budget. BM25 statistics are
recomputed per bank. Original entries are judged using Gaussian renders at
their stored cameras. All 1,329 output pairs have GPT judgments;
original-only predictions are reused unchanged.

\begin{table}[t]
\centering
\caption{\textbf{Adding grid views to original images.}
Historical fusion compares 18,570 original training images (O) with O plus 30,522
Gaussian grid renders (O+G). Each query searches its own scene.
GPT-confirmed success (\%); uncertain labels remain in the denominator.
Bold marks higher confirmed success within each row and metric, including ties.}
\label{tab:view_augmentation}
\footnotesize\setlength{\tabcolsep}{3pt}
\begin{tabular}{@{}lrcccc@{}}\toprule
Description source & $n$ & \multicolumn{2}{c}{Top-1 $\uparrow$} & \multicolumn{2}{c}{Top-5 $\uparrow$} \\
\cmidrule(lr){3-4}\cmidrule(lr){5-6}
 & & O & O+G & O & O+G \\ \midrule
All descriptions & 180 & 57.8 & \textbf{60.0} & 73.3 & \textbf{74.4} \\
Original-image source & 74 & \textbf{52.7} & \textbf{52.7} & \textbf{71.6} & 70.3 \\
Novel-render source & 106 & 61.3 & \textbf{65.1} & 74.5 & \textbf{77.4} \\
\bottomrule\end{tabular}
\end{table}

Table~\ref{tab:view_augmentation} shows four additional confirmed Top-1 and
two Top-5 successes across 180 descriptions. Gains occur among novel-render
descriptions; the 74 original-image descriptions have no Top-1 gain and one
fewer Top-5 success. All paired label-uncertainty bounds include zero, and
source groups differ in scene proportions. Thus, the 2.64-fold bank expansion
has no established consistent benefit; it also does not test the current BM25
interface's augmentation gain. Supplementary Sec.~\ref{sec:view_bank_supp}
reports same-pose, grid-only, density, and fixed-size replacement controls.

\subsection{Query Latency}
\label{sec:exp_efficiency}

QGVI-T fusion takes 0.020/0.063\,s median/p95 including CPU encoding,
retrieval, and place selection (Table~\ref{tab:efficiency}). Text timers and
hardware differ; ranking-only CLIP/BM25/MiniLM timings are excluded.

RGB timing uses four queries per scene, three resident repeats, and an
RTX PRO 6000 Blackwell GPU. Timers include decoding, preprocessing,
localization, acceptance, and device synchronization, excluding map/model
loading. QGVI-I takes 0.240/0.450\,s median/p95. Its photometric ablation,
including the final gate, has not been timed; native LoGS timing excludes that gate.
Supplementary Sec.~\ref{sec:supp_additional_latency} gives per-scene times,
timer boundaries, and repeatability checks.

\begin{table}[t]
  \centering
  \caption{\textbf{Measured query latency.}
  Median/nearest-rank 95th percentile in seconds, pooled over all nine scenes.
  Text: 180 queries (540 observations per CityLoC arm); RGB: 36 fixed
  queries, three repeats. Empty outputs remain in the denominators.
  QGVI-T (fusion) denotes historical CPU retrieval without online VLM calls. Text
  timer boundaries differ; bold marks the fastest text latency and compares
  qualified RGB latencies.}
  \label{tab:efficiency}
  \footnotesize
  \setlength{\tabcolsep}{3pt}
  \begin{tabular}{@{}lc@{}}
    \toprule
    Method & \shortstack{Query time (s) $\downarrow$\\Median / p95} \\
    \midrule
    CityLoC w/o refinement & 3.376 / 8.389 \\
CityLoC with refinement & 171.720 / 429.821 \\
QGVI-T (fusion) & \textbf{0.020} / \textbf{0.063} \\
\midrule
LoGS & 0.436 / 2.293 \\
SplatHLoc & 0.510 / 0.847 \\
QGVI-I & \textbf{0.240} / \textbf{0.450} \\
\bottomrule

  \end{tabular}
\end{table}

\section{Discussion and Limitations}
\label{sec:discussion}

\paragraph{Refinement and acceptance.}
Correct-pose retention is empirical: accurate poses below 40 distinct
inliers may be rejected. Geometric support provides no distribution-free
risk bound; the calibrated distinct-count operating point does not
minimize integrated risk across thresholds.

\paragraph{Text success and abstention.}
Historical fusion verification's 45/180 empty returns and QGVI-A's 49/180
nonreturns do not prove absence of a satisfying view. Rejecting cross-scene
RGB distractors does not establish robustness to arbitrary appearance,
sensor, or content changes.

\paragraph{Retrieval and view coverage.}
Optional grid expansion enlarges the bank 2.64-fold without an established
consistent benefit. BM25 and agentic runs retain this bank. Label bounds
are not statistical significance tests.

\paragraph{Evaluation scope.}
Camera moves need not be traversable; predicate checks cannot establish
scene-wide answerability. QGVI-A has more correct target-view returns but lower
answer precision than Retrieve + Select (92.6--93.0\% versus 98.5\%).
Information, call budgets, and transports differ, precluding a controlled latency
ranking. Source-preserving static verification with matched information and
call budgets remains untested.
Querying a Gaussian asset does not establish superiority over other scene representations.
Model-judge agreement measures consistency rather than human correctness.
DA3 supplies a fixed scale conversion where source metric scales are
unavailable, not surveyed ground truth. Reconstruction reuse and scale
calibration are specified in Supplementary Sec.~\ref{sec:supp_protocol}.

\section{Conclusion}
\label{sec:conclusion}

Queryable Gaussian View Index links QGVI-T's semantic view index and QGVI-I
localization through QGVI-A's tool calls within supplied Gaussian assets.
Our nine-scene development benchmark spans text, RGB, and target-view queries.
QGVI-A reduces wrong text returns relative to BM25 at lower coverage.
GPT judges 226/250 target answers correct, versus 67 for Retrieve + Select and
44 for Fixed Search + Verify; QGVI-A makes more errors, with differing
information and budgets. QGVI-I preserves every correct geometric pose while
rejecting most distractors. Controls support a geometric default, without
established grid-augmentation or photometric-refinement accuracy gains.
Correctness remains task-specific; semantic judgments remain model assessments.

\FloatBarrier
\label{page:main_end}
\label{page:references_start}
{\fontsize{9}{10.5}\selectfont
\setlength{\bibsep}{0pt}

}
\label{page:references_end}
\clearpage
\appendix
\raggedbottom
\makeatletter
\setlength{\@fptop}{0pt}
\setlength{\@fpsep}{18pt}
\setlength{\@fpbot}{0pt plus 1fil}
\setlength{\@dblfptop}{0pt}
\setlength{\@dblfpsep}{18pt}
\setlength{\@dblfpbot}{0pt plus 1fil}
\makeatother
\setcounter{table}{0}
\renewcommand{\thetable}{S\arabic{table}}
\renewcommand{\theHtable}{supp.\arabic{table}}
\section{Supplementary Material}
\label{sec:supp_historical}

\subsection{Historical Gaussian-Depth Pipeline}
The historical Gaussian-depth pipeline (v3) localizes images using
Gaussian-rendered RGB-D references and geometric verification. It differs
from QGVI-I in retrieval, local matching, 3D geometry, and verification.
Table~\ref{tab:historical_rgb} compares these complete designs;
Table~\ref{tab:selective_pose} isolates the QGVI-I acceptance gate at
fixed geometric initializer poses.

The historical pipeline uses SALAD~\cite{izquierdo2024salad} retrieval
with four diverse starting views,
RootSIFT~\cite{arandjelovic2012rootsift} correspondences based on
SIFT~\cite{lowe2004sift}, geometric pose estimation, and verification against
a newly rendered Gaussian RGB-D view. The query is resized to a maximum
side of 512 pixels before SALAD's standard 322-pixel input transform.
The reference bank, 512-pixel local features, solver, gates, and render budgets
remain fixed. CLIP embeddings are retained for diagnostics and do not
determine geometric acceptance. This variant uses the original
Horizon-GS~\cite{jiang2025horizongs} checkpoints and frozen native reference banks.

Each reference is back-projected with its own intrinsics. Pose estimation
and verification renders use the query intrinsics, so retrieval does not
require the reference and query cameras to have matching calibration.

The historical pipeline achieves 81.1\% strict scene-macro pose success,
accepts no out-of-map queries, and rejects 381 in-map queries (20.7\%
pooled; 16.0\% scene mean). It also accepts 45 inaccurate poses; these
localization failures are distinct from false rejects.
QGVI-I has 97.3\% strict scene-macro pose success, 0.707\% pooled false
acceptance, and 1.142\% pooled false rejection on the same queries.
The historical pipeline rejects more in-map queries while accepting no
out-of-map query in this cohort. Neither result establishes a universal
rejection guarantee.

\begin{table*}[t]
  \centering
  \caption{\textbf{Historical Gaussian-depth pipeline and QGVI-I.}
  Both use the same nine-scene development queries and fixed scale factors.
  Pose success uses all in-map queries; false acceptance uses all out-of-map
  queries; false rejection uses all in-map queries. Rates are percentages;
  scene means weight scenes equally. $^\dagger$ DA3-estimated scale.
  This comparison changes several components jointly and does not isolate
  the acceptance gate. Bold marks best values within each scene and metric,
  including ties.}
  \label{tab:historical_rgb}
  \fontsize{8}{9.5}\selectfont
  \setlength{\tabcolsep}{2pt}
  \begin{tabular*}{\textwidth}{@{\extracolsep{\fill}}p{75pt}p{72pt}*{10}{c}@{}}
    \toprule
    Method & Metric (\%) & \shortstack{Block\\Small} & \shortstack{Block\\A} & NYC & SF & Citysample & Colosseum & Elvenruin & Park & Road & \shortstack{Scene\\mean} \\
    \midrule
    In-map queries & $n$ & 139 & 602 & 180 & 174 & 142 & 126 & 118 & 228 & 130 & --- \\
    Out-of-map queries & $n$ & 139 & 602 & 180 & 174 & 142 & 126 & 118 & 228 & 130 & --- \\
    Scale & m / native unit & 25 & 25 & 1 & 1 & 24.777$^\dagger$ & 12.287$^\dagger$ & 14.135$^\dagger$ & 1.040$^\dagger$ & 1.339$^\dagger$ & --- \\
    \midrule
    \multirow{4}{*}{\shortstack[l]{Gaussian-depth\\pipeline (v3)}} & Pose 1\,m/2$^\circ$ $\uparrow$ & 63.3 & 60.6 & \textbf{100.0} & \textbf{100.0} & 88.0 & 64.3 & 94.9 & 82.9 & 76.2 & 81.1 \\
     & Pose 5\,m/5$^\circ$ $\uparrow$ & 64.0 & 62.0 & \textbf{100.0} & \textbf{100.0} & 95.8 & 64.3 & 97.5 & 84.2 & 78.5 & 82.9 \\
     & False accept $\downarrow$ & \textbf{0.0} & \textbf{0.0} & \textbf{0.0} & \textbf{0.0} & \textbf{0.0} & \textbf{0.0} & \textbf{0.0} & \textbf{0.0} & \textbf{0.0} & \textbf{0.0} \\
     & False reject $\downarrow$ & 36.0 & 37.7 & \textbf{0.0} & \textbf{0.0} & 4.2 & 29.4 & 2.5 & 14.0 & 20.0 & 16.0 \\
    \midrule
    \multirow{4}{*}{\textbf{QGVI-I}} & Pose 1\,m/2$^\circ$ $\uparrow$ & \textbf{90.6} & \textbf{93.4} & \textbf{100.0} & \textbf{100.0} & \textbf{100.0} & \textbf{91.3} & \textbf{100.0} & \textbf{100.0} & \textbf{100.0} & \textbf{97.3} \\
     & Pose 5\,m/5$^\circ$ $\uparrow$ & \textbf{92.1} & \textbf{93.5} & \textbf{100.0} & \textbf{100.0} & \textbf{100.0} & \textbf{91.3} & \textbf{100.0} & \textbf{100.0} & \textbf{100.0} & \textbf{97.4} \\
     & False accept $\downarrow$ & \textbf{0.000} & 0.332 & 1.111 & 4.023 & 1.408 & \textbf{0.000} & \textbf{0.000} & \textbf{0.000} & \textbf{0.000} & 0.764 \\
     & False reject $\downarrow$ & \textbf{3.597} & \textbf{2.492} & \textbf{0.000} & \textbf{0.000} & \textbf{0.000} & \textbf{0.794} & \textbf{0.000} & \textbf{0.000} & \textbf{0.000} & \textbf{0.765} \\
    \bottomrule
  \end{tabular*}
\end{table*}

\subsection{Representation and Initialization Diagnostics}
\paragraph{Map representation.}
With the original maps, the historical pipeline accepts 1,458 in-map poses,
of which 1,413 satisfy 1\,m/2$^\circ$. Pose success is 76.8\% when pooled
over all 1,839 in-map queries and 81.1\% when scenes receive equal weight;
neither denominator excludes rejections. On identical query images and
reference camera poses, the earlier common-SH3 adaptation achieves 19/139
strict successes in Block Small and 9/602 in Block A, compared with 88/139
and 365/602 for the original-map system. Reference renders, depth, retrieval
encoder and selections, and camera adapter change together, preventing
attribution to Gaussian map quality alone. The original maps follow the
benchmark configuration, but the transition was evaluated after inspecting
earlier outcomes. This is therefore a development comparison; the
common-SH3 scores remain separate from Table~\ref{tab:historical_rgb}.

\paragraph{Query composition.}
The earlier 96.0\% result uses a separate set of 200 rendered Block A
queries, equally split between aerial and street views, inside a supported
256-patch atlas with 24,576 references. That cohort is distinct from the
602 Block A positives in the nine-scene benchmark. Using the same atlas
on these 602 queries yields 97 strict successes
(16.1\%), compared with 365 (60.6\%) for the historical SALAD-initialized
system using the original representative bank. This comparison changes the
initializer as well as reference coverage. The 96.0\% estimate therefore
characterizes the supported-atlas setting; it does not describe performance
on the nine-scene pool.

A source-image comparison identifies 374 queries shared with the
earlier Block A benchmark. Source pixels, camera rotations, metre-converted
translations and resized intrinsics agree for every matched image. The
earlier CLIP-initialized native-v3 run has 113 strict successes (30.2\%);
the historical SALAD-initialized system has 231 (61.8\%), versus 306 (81.8\%)
for the archived Gaussian-only SALAD control. Cohort membership and pose
labels alone therefore do not explain the remaining gap. The control changes
retrieval, reference views, local feature resolution to 960 pixels, and
verification; it is excluded from the main method comparison.

\paragraph{Camera calibration.}
An earlier same-intrinsics restriction excluded 16 of Park's 228 in-map
queries and limited some Road queries to 12 of 2,669 references. Removing
that restriction and rendering at query calibration increases strict success
from 80 to 109 queries in Park and from 47 to 75 in Road. Colosseum and
Elvenruin each lose one success; the other scenes are unchanged. The restricted
adapter rejected 1,518 negatives without matching, whereas the corrected
CLIP-initialized system matches all 1,839. With CLIP, Block A retains 172 strict successes;
of its 2,408 positive-query initial hypotheses, 1,324 have insufficient matches
and 590 fail RANSAC. These counts locate the dominant loss at initialization;
they do not isolate retrieval from appearance or reference coverage.

\paragraph{Initializer selection.}
A paired comparison on all 988 validation queries holds the
reference bank, local geometry, budget of four starting views, and acceptance gates fixed.
Replacing CLIP ranking with SALAD improves strict scene-macro pose success
from 62.6\% to 80.0\%, with zero accepts among 494 negatives in each arm.
The selection rule requires higher strict success, no increase in either
scene-macro false acceptance or total negative accepts, and no additional
execution failures. The SALAD recipe was fixed before the full test run and
was not retuned on its outcomes. On the complete test cohort, strict success
increases from 62.7\% to 81.1\%, including 172 to 365 strict successes in
Block A. Both query pools retain historical development exposure.

\paragraph{Unresolved geometric conventions.}
The source audit identifies an unexpressed half-pixel conversion between
Gaussian raster coordinates and OpenCV feature-depth lifting. A conditional
CPU consistency replay changes initial poses without changing gates; the
source-image pixel convention and final localization effect remain unresolved.
No calibration correction is applied to the historical pipeline in
Table~\ref{tab:historical_rgb}. Repeated
SIFT orientations can also count one physical landmark more than once in
correspondence support. These limitations remain separate from the measured
initializer comparison.

\subsection{Reference-View Coverage}
\label{sec:exp_bank}

Orientation coverage is evaluated in the historical CLIP-initialized
common-SH3 adaptation using three reference banks: the baseline grid, an
equal-size bank replacing selected views with upward-facing views at the
same grid positions, and a bank containing all additional upward-facing views. The equal-size comparison holds the
reference count fixed and retains every query. On Block Small, both modified
banks add three successful queries at each pose threshold: success at
1\,m/2$^\circ$ increases from 19/139 (13.7\%) to 22/139 (15.8\%). Full
expansion adds no further successes over equal-size replacement. Block A
remains at 9/602 (1.5\%) for both thresholds under all three banks. None of
the six runs accepts an out-of-map distractor. These results concern the two
exposed development scenes and the selected distractor pools.

\begin{table*}[t]
  \centering
  \caption{\textbf{Auxiliary common-map reference-bank ablation for RGB localization.}
  These six historical CLIP-initialized common-SH3 runs share the original query IDs and one
  fixed map; they are separate from the historical native-map comparison in
  Table~\ref{tab:historical_rgb}. Denominators are 139 in-map and 139 out-of-map
  queries for Block Small, and 602 of each for Block A. Query-time quantiles
  cover all queries, including rejected ones, from encoding through retrieval,
  pose search, verification when reached, and result transfer; model/cache
  startup and queue waiting are excluded. Quantiles use linear interpolation
  between ordered query times. Within each scene, bold marks higher pose
  success and lower false acceptance or time, including ties. Median and
  95th-percentile times are compared separately.}
  \label{tab:bank}
  \small
  \setlength{\tabcolsep}{4pt}
  \begin{tabular}{llrcccc}
    \toprule
    Scene & Reference bank & Views & \shortstack{Pose success $\uparrow$\\1\,m/2$^\circ$ (\%)} & \shortstack{Pose success $\uparrow$\\5\,m/5$^\circ$ (\%)} & \shortstack{Out-of-map\\accept (\%) $\downarrow$} & \shortstack{Query time (s) $\downarrow$\\Median / 95th percentile} \\
    \midrule
    Block Small & Baseline & 1,886 & 13.7 & 14.4 & \textbf{0.0} & 0.117 / 0.588 \\
    & Equal-size replacement & 1,886 & \textbf{15.8} & \textbf{16.5} & \textbf{0.0} & \textbf{0.109} / \textbf{0.587} \\
    & Full expansion & 1,945 & \textbf{15.8} & \textbf{16.5} & \textbf{0.0} & \textbf{0.109} / 0.591 \\
    \midrule
    Block A & Baseline & 13,094 & \textbf{1.5} & \textbf{1.5} & \textbf{0.0} & 0.125 / 0.221 \\
    & Equal-size replacement & 13,094 & \textbf{1.5} & \textbf{1.5} & \textbf{0.0} & \textbf{0.124} / \textbf{0.209} \\
    & Full expansion & 14,347 & \textbf{1.5} & \textbf{1.5} & \textbf{0.0} & 0.126 / 0.214 \\
    \bottomrule
  \end{tabular}
\end{table*}

\subsection{Historical Retrieval Measurements}
For the historical pipeline, retrieval timing includes reading, decoding, and resizing
the query, SALAD and retained CLIP encoding, and SALAD ranking with selection
of four starting views. GPU synchronization follows encoding. Model/index
loading, pose search, render verification, and output writing are excluded;
diagnostic descriptor writes between timed stages are also excluded. These
measurements characterize the historical initialization stages, not
optimized production latency or QGVI-I.
Table~\ref{tab:historical_retrieval} also preserves CLIP-based timings from
separate camera-support configurations and the 200-query Block A atlas.
Their cohorts and configurations remain distinct.

\begin{table*}[t]
  \centering
  \caption{\textbf{Historical RGB retrieval-stage latency.}
  Times are milliseconds; quantiles use linear interpolation and include
  rejected queries. (a) Historical SALAD-initialized geometric-v3, including
  retained CLIP encoding. (b) Separate CLIP-based configurations and cohorts.
  Timings are descriptive and are not ranked across scenes or cohorts.}
  \label{tab:historical_retrieval}
  \small
  \textbf{(a) Nine-scene historical geometric-v3}\par\smallskip
  \begin{tabular}{lrrr}
    \toprule
    Scene & Queries & Median (ms) $\downarrow$ & 95th percentile (ms) $\downarrow$ \\
    \midrule
    Block Small & 278 & 53.184 & 72.137 \\
    Block A & 1,204 & 101.922 & 122.144 \\
    NYC & 360 & 50.939 & 73.183 \\
    SF & 348 & 59.176 & 86.112 \\
    Citysample & 284 & 54.040 & 59.646 \\
    Colosseum & 252 & 44.233 & 59.522 \\
    Elvenruin & 236 & 46.095 & 62.974 \\
    Park & 456 & 59.687 & 86.181 \\
    Road & 260 & 59.922 & 79.409 \\
    \midrule
    Pooled & 3,678 & 61.076 & 120.157 \\
    \bottomrule
  \end{tabular}
  \par\medskip
  \textbf{(b) Other historical configurations}\par\smallskip
  \begin{tabular}{llrrr}
    \toprule
    Benchmark & Initializer & Queries & Median (ms) $\downarrow$ & 95th percentile (ms) $\downarrow$ \\
    \midrule
    Nine-scene RGB, cross-camera & CLIP (historical) & 3,678 & 34.046 & 75.635 \\
    Nine-scene RGB, camera-filtered & CLIP (historical) & 3,678 & 30.529 & 89.236 \\
    Block A supported atlas & CLIP (historical) & 200 & 43.817 & 47.198 \\
    \bottomrule
  \end{tabular}
\end{table*}

\section{Implementation Details}
\label{sec:implementation_details}

\paragraph{Posed view bank.}
The historical bank has 30,522 renders at 2,330 places. Centers form a
deterministic lattice near training cameras, borrowing their intrinsics
and interpolating nearby camera heights. A place identifier combines scene,
camera domain, and center hash, grouping orientations at one center;
it is neither a semantic landmark nor a radius cluster. Original-only and
original-plus-grid controls isolate augmentation (Sec.~\ref{sec:exp_view_bank});
Supplementary Sec.~\ref{sec:view_bank_supp} reports density and replacement.

\paragraph{QGVI-T caption search.}
Retrieval indexes caption summaries, objects, attributes, relations, and
visible text, excluding uncertainty. BM25 uses $k_1=1.2$, $b=0.75$.
The current interface adds 18,570 originals to the rendered bank. It returns
up to six positive-score candidates, deduplicating image hashes and full
poses rounded to six decimals; view IDs break ties. No embeddings are loaded.

\paragraph{Historical fusion and verification.}
Earlier comparisons pool maximum MiniLM sentence cosine similarity per
view~\cite{wang2020minilm,reimers2019sentencebert} and fuse semantic and BM25
ranks~\cite{cormack2009rrf}: $s_i=(60+r_i^{\mathrm{sem}})^{-1}
+\mathbf{1}[b_i>0](60+r_i^{\mathrm{lex}})^{-1}$, with one-based ranks and
BM25 score $b_i$. View IDs break ties. Among 64 candidates, first occurrences
of place identifiers precede repeats. The first 20 form a shortlist returning
up to five distinct places with unchanged poses, without visual verification.
Offline captioning and the optional verification ablation use DeepSeek Flash
at temperature zero and maximum image side 1,280 pixels. Verification checks
shortlisted images without captions and ranks eligible views by relevance.

\paragraph{QGVI-I local features.}
The LoGS geometric initializer uses at most 4,096 SuperPoint keypoints,
a 1,024-pixel maximum image side, non-maximum suppression radius three,
and outdoor SuperGlue weights with five Sinkhorn iterations. QGVI-I
returns the geometric initializer and tests distinct-keypoint support
there; it does not load a Gaussian map at query time. The optional
QGVI-I + photo. ablation runs the native LoGS photometric optimizer
for up to 20 steps and tests support at the refined pose.

\paragraph{Acceptance calibration.}
The calibration set contains 245 in-map and 245 out-of-map queries from
Block A and SF. QGVI-I uses a 343-combination search over distinct counts,
inlier fraction, and image coverage. It selects successful PnP and at
least 40 distinct inlier query keypoints at the initializer, with fraction
and coverage thresholds zero. The rule is fixed across all nine test
scenes. Matched count controls use $\{0,4,8,12,20,40,80\}$ and retain
every validation pose correct at either 1\,m/2$^\circ$ or 5\,m/5$^\circ$.
Selection maximizes out-of-map rejection, then in-map acceptance; ties
favor fewer active gates and smaller thresholds. The optional
QGVI-I + photo. ablation requires native final-pose eligibility and at
least 40 distinct inlier query keypoints with positive depth and
reprojection error at most 12 pixels at the refined pose.

\section{Additional Experimental Protocol and Diagnostics}
\label{sec:supp_protocol}

\paragraph{Source split and text cohorts.}
The localization source split contains 18,570 training and 1,309
evaluation images with no exact file overlap. Some evaluation observations
contributed to reconstruction; the evaluation therefore does not test
generalization to unseen reconstruction observations. In the 180-query
text cohort, source-image review labels 160 descriptions fully supported,
14 partially supported, and six unsupported. All remain in the denominator.
A description unsupported by its source image may still match another view.
Development exposure includes 96 previously used source images and one
repeated normalized description. The reported cohorts are development
evaluations; no independent final evaluation has been conducted.
The 180- and 135-query text cohorts use RGB test-source images; the
176-query cohort uses RGB validation-source images. The 135 descriptions
are synthetic; the other two cohorts are DeepSeek-authored. Within each
comparison, methods share query IDs, source splits, and target scenes.
The overall text-ablation and breakdown rows pool the 176 query IDs.

\paragraph{RGB cohorts and cached evaluation.}
The 1,839 test positives comprise 480 aerial and 1,359 street views.
Block A contributes 501 source images and 101 renders, versus 13 and 113
in Colosseum. The separate 988-query validation pool contains 280
source-image positives, 214 rendered positives, and 494 negatives.
Out-of-map queries have no pose error and enter false-acceptance
evaluation, not pose-success denominators. An always-accepting method has
100\% false acceptance regardless of in-map accuracy. Initialization
failures remain in the reported pose-success denominators.
QGVI-I evaluation reuses frozen initializer outputs and correspondences;
fresh runs on four queries reproduce those cached outputs. The optional
QGVI-I + photo. ablation uses frozen native LoGS refined poses and
recomputes distinct support from saved correspondences. The GSplatLoc
appearance diagnostic is excluded from scored comparisons.

The 13 QGVI-I false accepts occur in Block A (2), NYC (2), SF (7), and
Citysample (2). QGVI-I + photo. has 10: two in Block A, one in NYC,
five in SF, and two in Citysample. Both reject the same 21 in-map queries
in Block Small (5), Block A (15), and
Colosseum (1). Map-training coverage also differs: the released LoGS schedule
loads 3,100 of Park's 6,194 allowed cameras over 30,000 iterations;
the number of distinct cameras used in optimization is unrecorded.
All nine scenes are reported.

\paragraph{Scene-scale verification.}
The source-verified scales are 25 metres per model unit for Block Small
and Block A, and one for NYC and SF. The latter are checked against 351 paired camera
shifts with labelled 0.3\,m height differences and 774 source-to-checkpoint
and checkpoint-to-benchmark camera matches per scene. The other five
scene scales use a calibrated ratio of DA3 raw depth to Gaussian-rendered
expected camera-Z depth on frozen training street views. A conversion
constant is fitted using 13 Block Small views and its verified scale,
then applied unchanged to the median depth ratio in each target scene.
Every method uses the same frozen scene-wide factor; it is not fitted to
localization outcomes. Block A, held out from scale calibration, has a
scale error of 3.9\%. Calibration cameras retain their native coordinates,
so scale transfer does not require map rescaling. This procedure provides
a shared metric conversion for comparison, while the five DA3 estimates
remain unverified by independent physical measurements.

\paragraph{Historical text retrieval diagnostics.}
All 180 queries have joint judgments; scene-macro and pooled rates agree
after seed averaging. A one-query CityLoC repeatability check produced
different refined poses and rankings despite fixed inputs and seed;
no independent paired refinement effect has been measured.
On the 135-query cohort, fusion with verification produces three
Top-5-only gains and seven losses relative to caption retrieval, a net loss of four. Six
losses return no result; the seventh ranks its relevant view sixth.
All nine successful caption-retrieved views from these seven queries
were in the 20-view shortlist. These observations do not
support a tuning claim. For the 176-query
shortlists, caption and CLIP success-count bounds are 152--176 and
141--175; these are not full-bank recall estimates. Returned-view validity
is judged for the 135- and 180-query cohorts but unverified for the
176-query cohort. On the 180-query cohort, CityLoC refinement raises
seed-averaged Top-5 success from 30.7\% to 34.6\%, with 19.8\% no result.

\paragraph{Historical query timing boundaries.}
The earlier 176-query efficiency run uses PyTorch 2.9.1+cu128. CityLoC timings
include serialization and GPU synchronization, exclude model/map startup,
and use all queries with nearest-rank 95th percentiles. Peak PyTorch
allocated memory includes resident models/maps but excludes driver and
custom non-PyTorch allocations. The separate 135-query pilot uses
135 IDs across three seeds, including only one ID each in NYC and SF;
per-scene median times are 2.5--7.3\,s without refinement and
35.6--287.6\,s with it. Returning no result does not establish correct rejection.

That historical fusion + verification timing pass uses 3,562 HTTP attempts across
all 176 queries and retries (20.24 per query); two query IDs end in API-response failure.
The execution history contains four failed executions, including earlier
retries; this count differs from the two failed query IDs. Failed executions
enter neither latency quantiles nor retrieval no-result counts. Cached-response
diagnostics are excluded from online timing. Provider GPU usage and
memory are unknown.

\section{Additional Evaluation Protocols}
\label{sec:supp_additional}

The following evaluations separate recovery of the description's source
camera, semantic satisfaction, and query latency.
The 180-query development cohort and method outputs remain fixed.
QGVI-T here denotes the historical fusion variant, not the current BM25 interface.

\subsection{Text Source-Camera Recovery}
\label{sec:supp_source_pose}

Each description is linked to the camera that produced its authoring image.
For a predicted camera-to-world pose $(\hat R,\hat c)$ and source pose
$(R^\star,c^\star)$, translation error is
$s\|\hat c-c^\star\|_2$, where $s$ is the frozen scene scale;
rotation error is the angle of $\hat R^\top R^\star$.
Stored retrieval poses can be evaluated directly.
These errors measure source-camera recovery: another view may satisfy the
description without being close to its source camera.

Top-1 pose success requires both translation and rotation errors to meet
the specified threshold. Pose Recall@5 requires at least one of the first
five outputs to meet both thresholds; the same output must satisfy both.
Empty returns fail, and shorter lists receive no additional candidates.
Missing query records or invalid pose matrices invalidate an evaluation
rather than becoming successful abstentions.
All methods use the same query IDs, fixed ranking, and output budget.
CityLoC's three seeds are averaged within each query before scene averaging.

We report the nine-scene mean and a separate mean over the four scenes
with source-verified metre scales. The remaining five use the fixed
DA3 estimates disclosed in Table~\ref{tab:nine_scene_detail}.
Top-1 error distributions are conditional on a returned valid pose and
must be accompanied by their denominator and return rate.
An auxiliary full-bank oracle tests whether any stored reference pose
can satisfy the thresholds; it measures the discrete bank's coverage,
not the retrieval method's accuracy.
The bank covers 0/180 source cameras at 1\,m/2$^\circ$ and 40/180
at 5\,m/5$^\circ$. For 171 descriptions, no bank orientation is within
2$^\circ$ of the source, irrespective of position. Independent checks
trace all source poses to the original held-out or novel-view records
and reproduce the pose scores with a second rotation implementation.
This coverage bounds QGVI-T's discrete retrieval, not CityLoC's continuous
pose proposals. At the looser threshold, QGVI-T retrieves 6 of the 40
queries covered by the bank within its five outputs; the optional
verification variant retrieves 8.

\begin{table*}[t]
  \centering
  \caption{\textbf{Text source-camera recovery.}
  Historical fusion-based QGVI-T; same descriptions and scene-macro percentages after seed
  averaging. P1 is Top-1 pose success and R5 is Pose Recall@5.
  Error columns average conditional scene medians, with coverage reported
  alongside the completed results. CityLoC first averages errors over
  seeds returning a valid Top-1 pose within each query.
  The four-scene subset uses source-verified metre scales; five other
  scenes retain DA3 estimates. All scenes contribute conditional
  errors. In row order, error distributions contain 180, 160, 180,
  and 135 queries (80, 72, 80, and 63 in the four-scene subset);
  pose coverage averages seed-level returns.
  Bold marks highest success and lowest error within each
  cohort; coverage is descriptive.}
  \label{tab:source_pose_pending}
  \footnotesize
  \setlength{\tabcolsep}{5pt}
  \begin{tabular}{@{}lcccccc@{}}
    \toprule
    Method & \shortstack{P1 $\uparrow$\\1\,m/2$^\circ$} & \shortstack{R5 $\uparrow$\\1\,m/2$^\circ$} &
    \shortstack{P1 $\uparrow$\\5\,m/5$^\circ$} & \shortstack{R5 $\uparrow$\\5\,m/5$^\circ$} &
    \shortstack{Mean scene median $\downarrow$\\(m/$^\circ$)} & Pose coverage (\%) \\
    \midrule
    \multicolumn{7}{l}{All nine scenes, 180 queries} \\
    CityLoC w/o refinement & \textbf{0.00} & \textbf{0.00} & 0.19 & 0.19 & 86.30 / 69.99 & 100.00 \\
    CityLoC with refinement & \textbf{0.00} & \textbf{0.00} & 0.37 & 0.56 & 89.16 / 64.98 & 80.19 \\
    QGVI-T & \textbf{0.00} & \textbf{0.00} & \textbf{0.56} & 3.33 & 64.66 / \textbf{51.64} & 100.00 \\
    QGVI-T + verification & \textbf{0.00} & \textbf{0.00} & \textbf{0.56} & \textbf{4.44} & \textbf{47.08} / 55.26 & 75.00 \\
    \midrule
    \multicolumn{7}{l}{Four scenes with source-verified scales, 80 queries} \\
    CityLoC w/o refinement & \textbf{0.00} & \textbf{0.00} & 0.42 & 0.42 & 125.34 / 68.42 & 100.00 \\
    CityLoC with refinement & \textbf{0.00} & \textbf{0.00} & 0.83 & 0.83 & 132.09 / 62.06 & 80.42 \\
    QGVI-T & \textbf{0.00} & \textbf{0.00} & \textbf{1.25} & 6.25 & 90.57 / \textbf{51.12} & 100.00 \\
    QGVI-T + verification & \textbf{0.00} & \textbf{0.00} & \textbf{1.25} & \textbf{7.50} & \textbf{60.33} / 51.75 & 78.75 \\
    \bottomrule
  \end{tabular}
\end{table*}

\subsection{Blinded Cross-Model Semantic Assessment}
\label{sec:supp_human}

The historical CityLoC/QGVI-T comparison in Table~\ref{tab:historical_text_utility} and the
earlier 180- and 135-query tables retain completed Qwen/DeepSeek judgments.
Qwen3-VL-8B-Instruct (revision \texttt{0c351dd0}) assigns one of four
relevance labels: fully relevant, partially relevant, irrelevant, or
uncertain. Separate DeepSeek image-only judgments assign render validity.
Joint success requires full relevance and a valid image; uncertain labels
produce lower and upper bounds. The completed Kimi/GPT assessment is
reported separately in Table~\ref{tab:judge_accounting}; its labels do not
alter those historical Qwen/DeepSeek labels. The common first-answer
text comparison reuses compatible GPT judgments and adds reviews of the
frozen BM25 and QGVI-A outputs (Supplementary Sec.~\ref{sec:supp_selective_answers}).

The completed reassessment pools fixed Top-5 outputs from CityLoC with
and without refinement and QGVI-T with and without verification, across
all seeds on the 180-query cohort: 6,150 output occurrences
yield 5,808 distinct query--pixel pairs. Identical pairs are assessed once
per reviewer and their labels reused for every occurrence. Each reviewer
receives only the description and rendered image; method, rank, camera
pose, stored captions, and earlier judgments are withheld.

Reviewer A uses Kimi (requested model
\texttt{kimi-for-coding}, Kimi Code CLI 2.2.0, low thinking effort);
Reviewer B uses GPT-6.1 (requested model \texttt{gpt-6.1-sol}, Codex CLI
0.159.2, low reasoning effort). These are requested model aliases; backend
version identities are not independently verified. They are called independently with the
same rubric: relevance is supported, contradicted, or uncertain, and
validity is valid, invalid, or uncertain. Their joint success criterion
requires supported relevance and valid imagery. Their judgments and
disagreements are retained separately. Failed or invalid responses were
retried without replacing valid judgments. Both reviewers completed all
5,808 pairs (11,616 judgments); uncertain judgments remain unresolved.
This three-way relevance rubric differs from Qwen's
historical four-way rubric: a partially relevant judgment may correspond
to contradiction or uncertainty. Comparing full support across the two
rubrics therefore does not imply exact category agreement.

Top-1 and Top-5 success include all queries, including empty returns;
View-level precision pools all returned views, whereas the primary
accepted-answer precision in Sec.~\ref{sec:supp_selective_answers} uses only
the first answer per query. No-result rate measures empty outputs
without establishing whether a description is unanswerable. Full-corpus
recall is unavailable because the complete relevant-view set is unknown.
These are model-generated labels, and inter-model agreement alone does
not establish their correctness.

Table~\ref{tab:judge_accounting} reports separate Top-1 and Top-5 scores and
first-answer accounting for these four historical text arms. For $A$ returned answers,
$S$ confirmed successes, $W$ confirmed wrong answers, and $u$ unresolved answers,
precision lies in $[S/A,(S+u)/A]$ and wrong-return rate in
$[W/N,(W+u)/N]$. The utility interval is
$[(S-W-u)/N,(S+u-W)/N]$; abstention earns zero.
Both judges give QGVI-T higher Top-5 lower bounds than its verified variant;
their Top-1 lower-bound ordering differs. Overlapping intervals limit
ranking claims. These are repeated assessments of the same development
queries, not an independent test set.

\begin{table*}[t]
\centering
\caption{\textbf{Separate Kimi and GPT assessments of historical text methods.}
All 180 development queries from nine scenes; each CityLoC variant has three
seeds, giving $N=540$ query--seed observations. Coverage and all quality
rates are percentages. Top-1 success is correct-return rate $Y$; precision
$P$ conditions on returned answers, while wrong-return rate $E$ divides by
all queries. Net utility $100U_1$ uses per-query reward $+1$ for a correct
first answer, $-1$ for a wrong answer, and zero for abstention.
Ranges are unresolved-label bounds, not confidence intervals. Bold marks
the best conservative bound within each reviewer: lower for success,
precision and utility, upper for wrong returns. Overlap limits ranking
claims. Native acceptance policies are retained. Historical Qwen relevance
and DeepSeek validity results appear in Table~\ref{tab:historical_text_utility}(a).}
\label{tab:judge_accounting}
\textit{(a) Coverage, retrieval success, and first-answer utility}\par\smallskip
\begingroup\small\setlength{\tabcolsep}{4pt}
\renewcommand{\arraystretch}{1.12}
\begin{tabular}{@{}lrrrrrrrr@{}}\toprule
Method & $N$ & Coverage $C$ & \multicolumn{2}{c}{Top-1 success $Y$ $\uparrow$} & \multicolumn{2}{c}{Top-5 success $\uparrow$} & \multicolumn{2}{c}{Net utility $100U_1$ $\uparrow$} \\
\cmidrule(lr){4-5} \cmidrule(lr){6-7} \cmidrule(lr){8-9}
 &  &  & Kimi & GPT & Kimi & GPT & Kimi & GPT \\\midrule
CityLoC w/o refinement & 540 & 100.0 & $22.2\text{--}61.1$ & $26.7\text{--}66.5$ & $37.8\text{--}84.6$ & $40.6\text{--}82.0$ & $[-55.6,\,22.2]$ & $[-46.7,\,33.0]$ \\
CityLoC with refinement & 540 & 80.2 & $25.6\text{--}47.0$ & $29.1\text{--}49.4$ & $37.4\text{--}64.1$ & $39.6\text{--}63.0$ & $[-29.1,\,13.9]$ & $[-22.0,\,18.7]$ \\
QGVI-T & 180 & 100.0 & $52.2\text{--}73.9$ & $\mathbf{61.1}\text{--}77.8$ & $\mathbf{72.8}\text{--}93.3$ & $\mathbf{73.9}\text{--}91.7$ & $[4.4,\,47.8]$ & $[22.2,\,55.6]$ \\
QGVI-T + verification & 180 & 75.0 & $\mathbf{56.7}\text{--}68.9$ & $60.6\text{--}68.9$ & $67.2\text{--}74.4$ & $68.3\text{--}72.8$ & $[\mathbf{38.3},\,62.8]$ & $[\mathbf{46.1},\,62.8]$ \\
\bottomrule\end{tabular}\endgroup

\par\medskip
\textit{(b) First-answer precision and wrong-return rate}\par\smallskip
\begingroup\small\setlength{\tabcolsep}{6pt}
\renewcommand{\arraystretch}{1.12}
\begin{tabular}{@{}lrrrr@{}}\toprule
Method & \multicolumn{2}{c}{Answer precision $P$ $\uparrow$} & \multicolumn{2}{c}{Wrong returns $E$ $\downarrow$} \\
\cmidrule(lr){2-3} \cmidrule(lr){4-5}
 & Kimi & GPT & Kimi & GPT \\\midrule
CityLoC w/o refinement & $22.2\text{--}61.1$ & $26.7\text{--}66.5$ & $38.9\text{--}77.8$ & $33.5\text{--}73.3$ \\
CityLoC with refinement & $31.9\text{--}58.7$ & $36.3\text{--}61.7$ & $33.1\text{--}54.6$ & $30.7\text{--}51.1$ \\
QGVI-T & $52.2\text{--}73.9$ & $61.1\text{--}77.8$ & $26.1\text{--}47.8$ & $22.2\text{--}38.9$ \\
QGVI-T + verification & $\mathbf{75.6}\text{--}91.9$ & $\mathbf{80.7}\text{--}91.9$ & $6.1\text{--}\mathbf{18.3}$ & $6.1\text{--}\mathbf{14.4}$ \\
\bottomrule\end{tabular}\endgroup

\end{table*}

\subsection{Measured Query Latency}
\label{sec:supp_additional_latency}

We report query latency for seven historical variants on nine scenes.
These timings do not measure QGVI-A or the current BM25 interface.
Matched retrieval controls have ranking-only timings; QGVI-I + photo. and
calibrated baseline gates have no separate query timing. Timers exclude
model and reference loading. Cached-response replay is not a fresh API query
measurement.

\paragraph{Text queries.}
CityLoC with/without refinement and QGVI-T with/without verification use
the same 180 descriptions. CityLoC has
540 query--seed observations per arm. Without refinement, its median/p95
is 3.376/8.389\,s; with refinement, 171.720/429.821\,s, with 107 empty
outputs. Both arms render diffusion proposals for CLIP selection and
therefore require the inherited Gaussian map. Their timers include
candidate-image serialization and hashing, but exclude model/map loading.
QGVI-T fusion takes 0.020/0.063\,s, including diversity selection.
QGVI-T + verification takes 15.465/17.067\,s; all 180 queries complete, 45 return
no result, and the timer includes 3,614 HTTP attempts, network round
trips, and retries.
Its output lists differ from the frozen accuracy predictions on 73 queries,
so timing-pass outcomes are not substituted for accuracy results.
QGVI-T runs on the CPU; its optional verification variant adds a remote service.
The historical text runs were recorded on different dates without a
matched hardware protocol; the remote service hardware is unknown.

Table~\ref{tab:text_latency_scene} reports each scene separately. Each
CityLoC cell pools 20 queries over three training seeds; each QGVI-T
variant has 20 queries per cell. Quantiles are calculated from the individual query
times, not averaged across seed or scene summaries.

\begin{table*}[t]
  \centering
  \caption{\textbf{Text query latency by scene (seconds; $\downarrow$).}
  Median/nearest-rank p95 from the recorded query runs. CityLoC has 60
  observations per cell and each QGVI-T variant has 20. Empty outputs remain included.
  QGVI-T uses historical fusion; verification includes remote calls and retries.}
  \label{tab:text_latency_scene}
  \footnotesize
  \setlength{\tabcolsep}{4pt}
  \begin{tabular}{@{}lcccc@{}}
    \toprule
    Scene & CityLoC w/o refinement & CityLoC with refinement & QGVI-T & QGVI-T + verification \\
    \midrule
    Block Small & 4.776 / 5.915 & 145.724 / 609.644 & 0.021 / 0.029 & 15.457 / 16.844 \\
Block A & 8.244 / 10.336 & 182.853 / 700.906 & 0.062 / 0.086 & 15.648 / 17.970 \\
NYC & 3.297 / 3.693 & 198.926 / 327.830 & 0.018 / 0.022 & 15.462 / 16.845 \\
SF & 2.717 / 3.660 & 177.899 / 347.767 & 0.020 / 0.024 & 15.334 / 16.155 \\
Citysample & 3.113 / 3.772 & 85.713 / 239.108 & 0.021 / 0.024 & 15.380 / 16.164 \\
Colosseum & 2.559 / 3.081 & 170.950 / 226.452 & 0.012 / 0.015 & 15.836 / 16.915 \\
Elvenruin & 3.411 / 3.974 & 115.690 / 429.821 & 0.011 / 0.014 & 15.293 / 16.075 \\
Park & 3.527 / 4.066 & 180.399 / 332.218 & 0.029 / 0.037 & 16.092 / 18.325 \\
Road & 3.453 / 4.046 & 151.034 / 257.865 & 0.019 / 0.027 & 15.077 / 17.638 \\
\bottomrule

  \end{tabular}
\end{table*}

\begin{table}[t]
  \centering
  \caption{\textbf{RGB latency by scene (seconds; $\downarrow$).}
  Median/nearest-rank p95 of twelve resident observations per method and
  scene. Bold marks the lowest displayed median and p95
  across three qualified methods, including ties. QGVI-I is the geometric
  default; QGVI-I + photo. has no matched timing.}
  \label{tab:rgb_latency_scene}
  \footnotesize
  \setlength{\tabcolsep}{3pt}
  \begin{tabular}{@{}lccc@{}}
    \toprule
    Scene & LoGS & SplatHLoc & QGVI-I \\
    \midrule
    Block Small & 0.487 / 0.789 & 0.529 / 0.630 & \textbf{0.286} / \textbf{0.342} \\
Block A & 0.425 / 0.722 & 0.527 / \textbf{0.544} & \textbf{0.329} / 0.647 \\
NYC & 1.065 / 2.732 & 0.399 / 0.877 & \textbf{0.193} / \textbf{0.246} \\
SF & 0.365 / 0.776 & \textbf{0.107} / 0.857 & 0.232 / \textbf{0.300} \\
Citysample & 0.343 / 1.424 & 0.543 / 0.649 & \textbf{0.198} / \textbf{0.262} \\
Colosseum & 0.318 / 0.366 & \textbf{0.118} / 0.613 & 0.227 / \textbf{0.287} \\
Elvenruin & 0.564 / 2.305 & 0.530 / 0.658 & \textbf{0.222} / \textbf{0.334} \\
Park & 0.508 / 0.979 & 0.385 / 0.735 & \textbf{0.229} / \textbf{0.307} \\
Road & 0.459 / 0.554 & \textbf{0.108} / 0.588 & 0.259 / \textbf{0.453} \\
\bottomrule

  \end{tabular}
\end{table}

\paragraph{Fresh RGB timing.}
All three RGB methods use the same 36 query IDs, selected by a fixed hash
of public IDs before timing, and three resident repetitions per query.
Images, calibration, models, and reference assets retain their frozen
protocols. QGVI-I requires fresh features, retrieved references, matches,
poses, and gate decisions to reproduce its saved implementation outputs.
LoGS checks exact initializer outputs and canonical camera, image, mask,
and optimizer inputs before executing the unchanged native refinement.
SplatHLoc checks public input pixels and its unchanged native configuration.
For the two refined baselines, differences from historical poses are
retained as repeatability diagnostics.

LoGS's initial numerical check also failed when rerunning its frozen
native implementation. In an eight-query diagnostic over Block Small
and NYC, 7/48 direct native calls exceeded the previous saved-pose
tolerance. Identical visible forward outputs and losses produced
different pose gradients. Subsequent timing therefore qualifies native
source and exact external-input conformance; numerical variation is
reported separately, without increasing the tolerance or selecting
closer runs. This does not establish that the variation is harmless.
The failed checks and fixed accuracy outputs are preserved.

For LoGS, SplatHLoc, and QGVI-I, timing starts before image decoding and
ends after the pose and acceptance decision with device synchronization. Model/reference loading and
qualification checks are outside the query timer. The timing subset does
not estimate full-test accuracy or represent an unseen test. All local
RGB geometry runs use an NVIDIA RTX PRO 6000 Blackwell Workstation Edition.
LoGS retains input references and clones eight initial optimizer scalars
inside the timer; input hashing and canonical-input checks follow it.
PyTorch uses two CPU threads; native OpenCV threading is retained
(32 threads recorded for SplatHLoc), so this is not a two-core host limit.
Quantiles are computed from pooled raw observations, with nearest-rank
95th percentiles. With twelve observations per scene, this percentile is
the observed maximum; this small timing sample does not establish a
stable population tail.

\section{Shared Answer Accounting and Net Query Utility}
\label{sec:supp_selective_answers}

The query routes return answers whose correctness can be evaluated, but
their correctness predicates differ. For text, the first returned view must
be usable and satisfy every required semantic constraint. For RGB, the query
must be in-map and the returned pose must satisfy both geometric thresholds.
An accepted out-of-map image or an inaccurate in-map pose is incorrect.
For image-and-instruction queries, the output view must be usable and
satisfy every required target predicate, including its relation to the
source image; recovering the source camera alone is insufficient.
This shared accounting follows selective prediction~\cite{geifman2017selective};
it does not equate semantic satisfaction with geometric accuracy.

For $N$ queries, let $a_i=1$ when query $i$ receives a first answer and
$z_i=1$ when that answer is correct. Set $a_i z_i=0$ for an empty return.
With $A=\sum_i a_i$, $S=\sum_i a_i z_i$, and $W=A-S$, define
\begin{equation}
 C=\frac{A}{N},\qquad P=\frac{S}{A},\qquad
 Y=\frac{S}{N}=CP,\qquad E=\frac{W}{N}.
 \label{eq:selective_answers}
\end{equation}
Coverage $C$ measures how often the method answers; accepted-answer precision
$P$ measures how often those answers are correct; correct-return rate $Y$
measures how often a query receives a correct answer. Wrong-return rate
$E$ counts incorrect answers over all queries. Selective risk is $1-P$.
If $A=0$, precision and selective risk are undefined, while $C=Y=E=0$.
The identity uses fractions and pooled counts; it need not hold after
independently averaging scene-level $C$ and $P$.

\paragraph{Explicit costs for wrong answers.}
Cost-based evaluation makes the relative consequences of answering and
abstaining explicit~\cite{franc2023reject}. We assign reward $+1$ to a
correct first answer, reward $-\lambda$ to an incorrect first answer, and
zero to abstention:
\begin{equation}
 U_\lambda=\frac{S-\lambda W}{N}=Y-\lambda E,
 \qquad \lambda\geq0.
 \label{eq:query_utility}
\end{equation}
An all-reject system scores zero; a system returning only wrong answers
scores negatively for $\lambda>0$. The identities $U_0=Y$ and
$U_1=2Y-C$ show that utility adds an explicit preference over the existing
counts, rather than additional evidence. We report $100U_\lambda$ in net
reward units per 100 queries, not as an accuracy percentage. Values can
fall below $-100$ when $\lambda>1$.

The cost grid $\lambda\in\{0,0.5,1,2,5\}$ is a descriptive sensitivity
analysis specified after the recorded evaluations. It was not used to
calibrate gate thresholds. The default designation follows review of
these exposed development results. Equal reward and penalty
($\lambda=1$) provide an interpretable example; no deployment cost has
been established. This standard utility construction is an evaluation
choice, not a methodological contribution.

Table~\ref{tab:selective_answers} uses frozen localization outputs. Panel (a)
applies one GPT-low relevance/validity rubric to text; panel (c) retains
separate high-effort predicate judgments for image-and-instruction outputs. CityLoC contributes three equally weighted seed
runs per query. RGB uses a balanced in-map/out-of-map population and the
1\,m/2$^\circ$ threshold, including the declared estimated-scale scenes.
The maximum possible RGB correct-return rate is 50\% and its maximum
$U_\lambda$ is 0.5 in this mixture because out-of-map queries have no correct
pose. Rejection receives no positive reward. Consequently, text and RGB panels
must not be ranked against each other or pooled into one score. The
in-map pose-success rate and out-of-map rejection rate remain separate
diagnostics in Table~\ref{tab:nine_scene_detail}.

\paragraph{Common low-effort text assessment.}
All 180 descriptions are evaluated with the same GPT-6.1 Sol low-effort
relevance/validity prompt and label schema. Relevance and render usability
are judged from the unchanged query and output image, without method identity
or controller rationale. Historical labels are reused only when query,
image bytes, requested model, effort, prompt, and schema match. New reviews
use native Codex sessions; historical reviews used the Codex CLI with
ChatGPT subscription authentication. These transports and service dates differ, so matching the requested
configuration does not guarantee identical backend execution.
The first unverified BM25 result is recovered from the frozen QGVI-A prefix;
original-image cameras are evaluated through their cached Gaussian renders.
Every method retains all 180 descriptions (three seeds for CityLoC), including
nonreturns. Table~\ref{tab:gpt_low_text_all} reports all six methods; historical
fusion and verification are not omitted from the comparison.
No missing review is treated as an incorrect answer. An uncertain label in
either dimension leaves the joint result unresolved, even when the other
dimension is negative. These judgments remain intervals; they are not
resolved by selecting a judge.

\begin{table*}[t]
\centering
\caption{\textbf{Complete common-rubric text comparison.} All six frozen methods
use GPT-6.1 Sol, low effort, and the same relevance/validity rubric. CityLoC
has three seeds per description; other methods have one output list.
Ranges retain uncertain labels. Bold marks the best conservative endpoint
(lower for precision, correct returns and utility; upper for wrong returns),
including ties; this does not establish statistical superiority.}
\label{tab:gpt_low_text_all}
\small\setlength{\tabcolsep}{4pt}
\begin{tabular}{@{}lrrrrrr@{}}\toprule
Method & $N$ & $C$ (\%) & $P$ (\%) $\uparrow$ & $Y$ (\%) $\uparrow$ & $E$ (\%) $\downarrow$ & $100U_1$ $\uparrow$ \\
\midrule
CityLoC w/o refinement & 540 & 100.00 & 26.67--66.48 & 26.67--66.48 & 33.52--73.33 & -46.67--32.96 \\
CityLoC with refinement & 540 & 80.19 & 36.26--61.66 & 29.07--49.44 & 30.74--51.11 & -22.04--18.70 \\
QGVI-T & 180 & 100.00 & 60.00--70.56 & 60.00--70.56 & 29.44--40.00 & 20.00--41.11 \\
QGVI-A & 180 & 72.78 & \textbf{90.84--93.89} & \textbf{66.11--68.33} & \textbf{4.44--6.67} & \textbf{59.44--63.89} \\
Historical fusion & 180 & 100.00 & 61.11--77.78 & 61.11--77.78 & 22.22--38.89 & 22.22--55.56 \\
Fusion + verification & 180 & 75.00 & 80.74--91.85 & 60.56--68.89 & 6.11--14.44 & 46.11--62.78 \\
\bottomrule\end{tabular}
\end{table*}

\begin{table*}[t]
\centering
\caption{\textbf{Historical text assessment protocols.} These saved operating
points retain their original Qwen/DeepSeek or high-effort GPT judgments;
they are not the matched low-effort comparison in Table~\ref{tab:main}.
Bold is retained from the original within-panel comparison.}
\label{tab:historical_text_utility}
\small\setlength{\tabcolsep}{4pt}
\begin{tabular}{@{}lrrrrrr@{}}\toprule
Method & $N$ & $C$ (\%) & $P$ (\%) $\uparrow$ & $Y$ (\%) $\uparrow$ & $E$ (\%) $\downarrow$ & $100U_1$ $\uparrow$ \\
\midrule
\multicolumn{7}{l}{\textit{(a) Historical Qwen/DeepSeek rubric}} \\
CityLoC w/o refinement & 540 & 100.00 & 25.00 & 25.00 & 75.00 & -50.00 \\
CityLoC with refinement & 540 & 80.19 & 32.56 & 26.11 & 54.07 & -27.96 \\
QGVI-T (fusion) & 180 & 100.00 & 58.33 & 58.33 & 41.67 & 16.67 \\
Fusion + verification & 180 & 75.00 & \textbf{88.15} & \textbf{66.11} & \textbf{8.89} & \textbf{57.22} \\
\midrule
\multicolumn{7}{l}{\textit{(b) Earlier QGVI-A high-effort predicate rubric}} \\
QGVI-A & 180 & 72.78 & 95.42--97.71 & 69.44--71.11 & 1.67--3.33 & 66.11--69.44 \\
\bottomrule
\end{tabular}
\end{table*}

\paragraph{Agentic evaluation protocol.}
QGVI-A uses GPT-6.1 Sol, low effort. Prefix and continuation share limits of 12
decisions (including actions, transitions, and finish) and six render attempts;
failures consume these budgets. QGVI-T's BM25 interface searches original-image
and grid captions, returning up to six candidates without embeddings.
The RGB tool uses QGVI-I's geometric initializer and distinct-support
gate. These runs do not evaluate the optional photometric refinement
inside QGVI-A. Calls are query-dependent; an episode need not invoke both tools.
Camera-local moves and bounded pose proposals produce Gaussian
renders without traversability constraints. The prefix checks geometric
support for source poses, or rendered target predicates and source identity
when supplied.
The 250 image-and-instruction queries use 125 source images: 32 queries in
Block A, 22 in Elvenruin, and 28 in each other scene. Text and RGB retain
180 and 3,678 queries. RGB reuses the historical pixels, calibration,
membership, scales, and thresholds; newly executed poses need not be identical.

The earlier semantic assessment used a fresh high-effort request from the same model
family, receiving the instruction, required predicates, and final output
render; image-and-instruction reviews also receive the source image.
Review images have a maximum side of 512 pixels. The reviewer
does not receive the controller's rationale. An unusable render or a violated
predicate is a failure; otherwise, any unresolved predicate yields an
uncertain judgment. These are model assessments, not human ground truth.
This high-effort protocol remains the target-view evaluation and a text
sensitivity result. The new common low-effort assessment supplies the matched
text comparison; the two protocols are not pooled.

\paragraph{Observed trade-offs.}
Under the historical Qwen/DeepSeek rubric, at $\lambda=1$ fusion yields
16.67 net reward units per 100
queries, versus $-50.00$/$-27.96$ for CityLoC. Verification yields 57.22
and dominates raw fusion for every $\lambda\geq0$: 119 versus 105 correct
first answers, and 16 versus 75 wrong answers. It sacrifices Top-5 success
and latency (Tables~\ref{tab:text_ablation},~\ref{tab:efficiency}).
Raw fusion's utility becomes negative for $\lambda>1.4$.

\begin{figure*}[t]
 \centering
 \includegraphics[width=\textwidth]{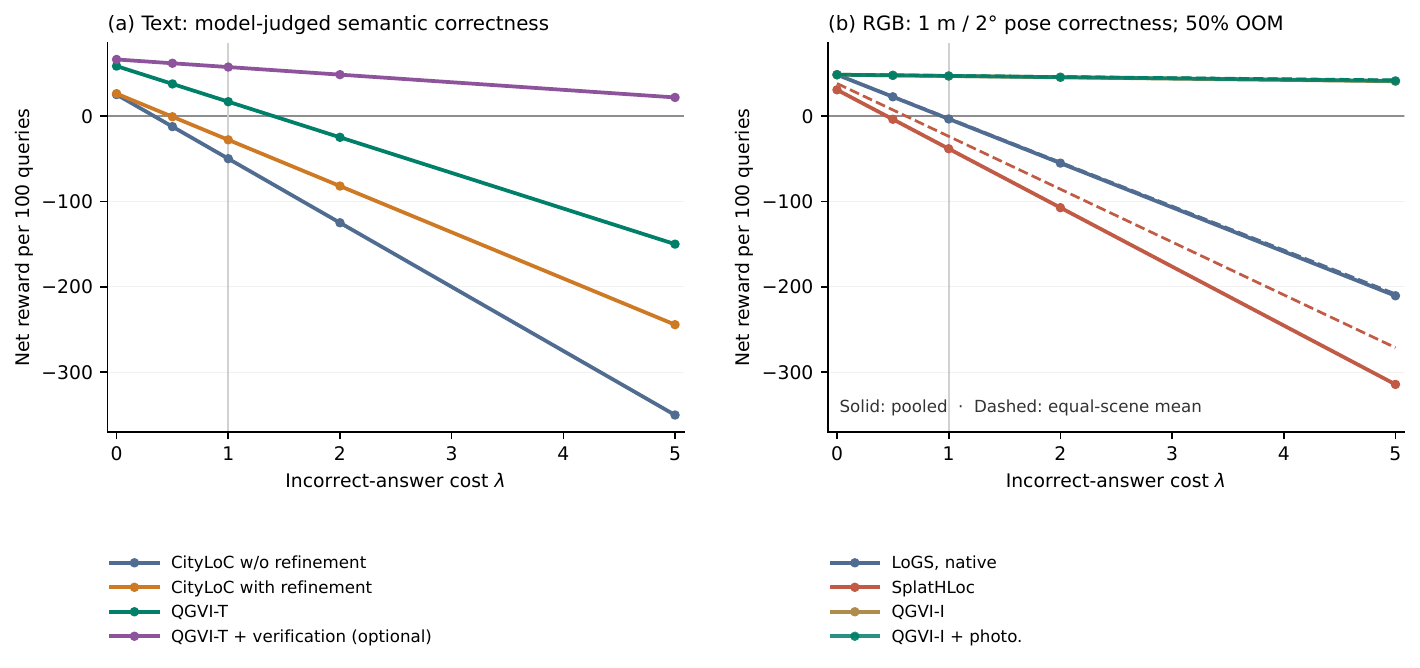}
 \caption{\textbf{Sensitivity to the cost of an incorrect first answer.}
 Correct answers earn $+1$, incorrect answers earn $-\lambda$, and
 abstentions earn zero. Curves connect the declared cost grid and are
 linear by definition. Solid curves use pooled query counts; dashed RGB
 curves weight scenes equally. Text aggregation schemes coincide.
 Text curves show historical fusion with Qwen/DeepSeek labels.
 The horizontal line is all-reject utility. The two modalities have
 different correctness definitions and query populations; their heights
 must not be compared. LoGS and SplatHLoc retain native acceptance; QGVI-I
 gates initializer poses, and QGVI-I + photo. gates refined final poses.
 Curves use frozen outputs, without new inference or threshold optimization.}
 \label{fig:utility_cost}
\end{figure*}

QGVI-I yields 46.74 per 100 RGB queries with 56 wrong returns: 43 inaccurate
in-map poses and 13 out-of-map accepts. QGVI-I + photo. yields 46.82 with
53 wrong returns (43 and ten); their utility curves nearly coincide.
Native LoGS/SplatHLoc yield $-3.48$/$-38.36$. Figure~\ref{fig:utility_cost}
shows pooled and equal-scene utility versus error cost. Both distinct-support
thresholds are 40; SplatHLoc's retention constraint selects its native count
gate (Table~\ref{tab:calibrated_gates}; Supplementary
Sec.~\ref{sec:calibrated_rgb_supp}).

Under the earlier high-effort protocol, QGVI-A returns 131 answers to
180 text queries: 125 are judged correct,
three incorrect, and three uncertain; 49 queries have no return.
On the 250 target-view queries, it returns 244 answers: 226 correct,
17 incorrect, and one uncertain. The six nonreturns include one invalid
model-output outcome. Table~\ref{tab:main}(c) now compares these fixed
outputs with two completed controls on the same target-view cohort;
their policies, budgets, and paired comparisons are detailed in
Sec.~\ref{sec:target_baselines}.
On RGB, QGVI-A returns 1,831 poses, including 1,773 strict successes,
46 inaccurate in-map poses, and 12 out-of-map accepts. Its net utility is
46.63, below QGVI-I's 46.74 and QGVI-I + photo.'s 46.82. The 20 rejected
in-map queries give a false-rejection rate of 1.088\%; the 12 accepted
out-of-map queries give a false-acceptance rate of 0.653\%.

\paragraph{Output budgets and unresolved labels.}
The shared primary definition uses one answer. Text Top-5 success remains a
separate fixed-budget measure: at least one of at most five returned views
must be correct. It does not imply that all five views are correct.
If $L$ accepted answers are labelled correct and $U$ have unresolved labels,
precision lies in $[L/A,(L+U)/A]$ and correct-return rate in
$[L/N,(L+U)/N]$. Utility lies in
\begin{equation}
 \left[\frac{(1+\lambda)L-\lambda A}{N},
 \frac{(1+\lambda)(L+U)-\lambda A}{N}\right].
\end{equation}
These are label-uncertainty bounds, not sampling intervals.
The matched low-effort text bounds retain every unknown judgment; no
reviewer responses are missing. The earlier high-effort text and target-view
protocols retain three and one uncertain judgments, respectively. CityLoC's 540 observations are three runs on 180 query
IDs; they are not 540 independent queries.
Missing reviewer responses do not become negative labels. Runtime failures
count as unsuccessful queries and are reported separately from deliberate
abstention; missing execution records invalidate an incomplete evaluation.

\paragraph{Acceptance and answerability.}
For a confidence score, thresholds should be selected on validation data
under a common retention or risk objective, then frozen for test evaluation.
Risk--coverage curves require a recorded score and an explicit threshold
policy; isolated binary decisions provide operating points, not a continuous
curve. The text cohort does not establish whether each description is
answerable anywhere in the scene. We therefore do not credit an empty text
return as correct rejection or report text false-accept/false-reject rates.
An unsupported source image does not imply that all other views fail.
Likewise, $Y$ is not answerability recall, so combining $P$ and $Y$ into
a harmonic mean would not produce a conventional F1 score. Full shared
risk--coverage curves cannot be recovered from the available operating
points: some native abstentions have no returned answer whose correctness
can be scored when hypothetically relaxing the policy.

\section{Target-View Retrieval and Fixed-Search Controls}
\label{sec:target_baselines}

We compare frozen QGVI-A outputs with two controls on the same 250
image-and-instruction queries and 125 source images. Required predicates,
source pixels, intrinsics, scene assets, and the final assessment rubric
are held fixed. Each source contributes two tasks. This is an exposed
development cohort, not an independent final test.

\paragraph{Control policies.}
\emph{Retrieve + Select} renders the initial six BM25 candidates using
the task camera calibration. In one GPT-6.1 Sol low-effort call, the model
sees the source and all available candidate renders and either selects
one supported existing view or reports insufficient evidence. It cannot
move a camera. \emph{Fixed Search + Verify} starts from the first BM25
candidate and inspects six precommitted poses: the anchor, yaw left/right
by $10^\circ$, then translate left/right/backward by a training-camera-derived
step. All offsets are relative to the anchor. The same low-effort model
checks one view at a time and stops at the first fully supported answer;
it cannot choose movements, inspect future views, or see prior judgments.
This deterministic control does not reproduce a separately published method.
Candidate cameras were frozen before new model outcomes; frozen adaptive
answers were not used to choose their movements. Exact renders are reused
only when camera, calibration, dimensions, renderer binding, and pixel
hashes agree; other candidates use the unchanged Gaussian renderer.

\paragraph{Budgets and assessment.}
Each control allows six render slots; invalid or out-of-bounds slots are
not replaced. Retrieve + Select uses 250 model calls and 1,500 slots in
total (means 1.000 and 6.000). Fixed Search + Verify uses 1,291 calls and
1,326 inspected slots (means 5.164 and 5.304). QGVI-A instead has shared
prefix/continuation limits of 12 decisions and six render attempts, with
failures consuming the budget. Decisions include actions, transitions,
and finish; they are not equivalent to model calls. The selected final
adaptive executions use 991 calls (mean 3.964, range 1--10), excluding
independent reviews and discarded recovery attempts.
The adaptive policy can
also inspect captions, reference photographs, history, and crops, and use
PnP or continuous pose proposals. The controls do not receive these same
observations or tools. Thus information access, feedback, stopping, and
computation differ: the comparison tests complete policies and does not
isolate navigation as a causal component.

Every returned image receives the same source-aware GPT-6.1 Sol high-effort
assessment used for the frozen adaptive output. The reviewer sees only
the source, final render, instruction, required predicates, and schema;
method, controller reasoning, and other judgments are withheld. This is
an independent call within the same model family, not human ground truth.
Exact matched historical assessments are reused and identical new review
inputs are deduplicated. Different session transports and service dates,
together with separately prepared shared-GPU renders, preclude a controlled
latency comparison; no runtime ranking follows from these records.

\paragraph{Counts and paired uncertainty.}
All 750 executions and all 362 returned-view assessments are complete.
Retrieve + Select returns 68 answers (67 correct, one wrong); Fixed Search
+ Verify returns 50 (44 correct, five wrong, one uncertain); QGVI-A returns
244 (226 correct, 17 wrong, one uncertain). Their no-return counts are
182, 200, and six, respectively. The adaptive six include one invalid
model-output failure; it is retained as a zero-utility nonreturn, separately
from deliberate abstention. No missing execution or assessment is scored
as a semantic failure.

Relative to QGVI-A, the controls' correct-return differences have
label-uncertainty bounds of $[-64.00,-63.60]$ and $[-73.20,-72.40]$
percentage points. Their utility differences are $[-57.60,-56.80]$ and
$[-68.80,-67.20]$ net reward units per 100 queries. These bounds enumerate
unresolved labels and are not confidence intervals. Separately, a paired
percentile bootstrap resamples the 125 source-image clusters, retaining
both queries per source (5,000 resamples; fixed seed 20261102). The
confirmed-success differences are $-63.60$ and $-72.80$ points, with 95\%
bootstrap intervals $[-70.80,-55.60]$ and $[-79.20,-66.00]$. This statistic
counts confirmed labels without imputing uncertain answers; its sampling
interval does not resolve label uncertainty or establish generalization
beyond the development cohort. Table~\ref{tab:target_baseline_detail}
shows the same accounting by scene and task family. Higher answer yield
does not imply uniformly higher precision: Retrieve + Select confirms
67 of 68 returned answers, while abstaining on 182 queries.

\begin{table*}[t]
\centering
\caption{\textbf{Target-view controls by scene and task family.}
All methods use the same queries within each row. $S/W/u$ are counts of
confirmed correct, wrong, and uncertain returns; the remaining $N-S-W-u$
queries have no return. Utility $100U_1$ uses every query. Ranges reflect
semantic-label uncertainty, not sampling intervals. Bold marks a utility
interval lying strictly above both alternatives. Policies and actual budgets
differ; these results do not isolate a navigation effect.}
\label{tab:target_baseline_detail}
\small
\setlength{\tabcolsep}{3pt}
\begin{tabular}{@{}lrcccccc@{}}
\toprule
& & \multicolumn{2}{c}{Retrieve + Select} & \multicolumn{2}{c}{Fixed Search + Verify} & \multicolumn{2}{c}{QGVI-A} \\
Group & $N$ & $S/W/u$ & $100U_1\uparrow$ & $S/W/u$ & $100U_1\uparrow$ & $S/W/u$ & $100U_1\uparrow$ \\
\midrule
Block Small & 28 & 6/0/0 & 21.43 & 2/0/0 & 7.14 & 27/1/0 & \textbf{92.86} \\
Block A & 32 & 5/0/0 & 15.62 & 3/1/0 & 6.25 & 31/1/0 & \textbf{93.75} \\
NYC & 28 & 7/0/0 & 25.00 & 3/1/0 & 7.14 & 23/2/1 & \textbf{71.43--78.57} \\
SF & 28 & 11/0/0 & 39.29 & 14/0/0 & 50.00 & 25/3/0 & \textbf{78.57} \\
Citysample & 28 & 7/0/0 & 25.00 & 5/0/0 & 17.86 & 25/1/0 & \textbf{85.71} \\
Colosseum & 28 & 14/0/0 & 50.00 & 10/2/0 & 28.57 & 25/2/0 & \textbf{82.14} \\
Elvenruin & 22 & 2/0/0 & 9.09 & 3/1/1 & 4.55--13.64 & 18/3/0 & \textbf{68.18} \\
Park & 28 & 11/1/0 & 35.71 & 2/0/0 & 7.14 & 25/3/0 & \textbf{78.57} \\
Road & 28 & 4/0/0 & 14.29 & 2/0/0 & 7.14 & 27/1/0 & \textbf{92.86} \\
\midrule
Multi-object composition & 80 & 12/0/0 & 15.00 & 6/1/0 & 6.25 & 69/9/0 & \textbf{75.00} \\
Relation disambiguation & 70 & 25/0/0 & 35.71 & 14/3/0 & 15.71 & 63/4/1 & \textbf{82.86--85.71} \\
Visibility / framing & 100 & 30/1/0 & 29.00 & 24/1/1 & 22.00--24.00 & 94/4/0 & \textbf{90.00} \\
\bottomrule
\end{tabular}
\end{table*}

\section{Nine-Scene Retrieval and Localization Details}
\label{sec:supp_nine_scene_detail}

Table~\ref{tab:nine_scene_detail} retains the per-scene text Top-1/Top-5
success, no-result rates, RGB pose success, false-accept/false-reject rates,
and scale factors. These diagnostics complement the pooled first-answer
accounting in main Table~\ref{tab:main}; text rows retain historical fusion
and Qwen/DeepSeek judgments, rather than the common GPT-low assessment.
The two panels use their original
query populations and metric denominators; scene means weight all nine
scenes equally.

\begin{table*}[tp]
  \centering
  \caption{\textbf{Text retrieval and image localization across all nine scenes.}
  Percentages; scene means weight all nine scenes equally. (a) Joint
  relevance/validity success under historical Qwen/DeepSeek judgments on
  180 development descriptions
  (20 per scene); CityLoC averages three seeds; QGVI-T uses historical fusion.
  Empty returns fail.
  (b) Pose success and false rejection use all in-map queries; false
  acceptance uses all out-of-map queries. LoGS and SplatHLoc use native
  maps. QGVI-I gates the LoGS geometric initializer without photometric
  refinement; the gate is calibrated on Block A/SF validation, then fixed.
  Baselines retain their
  acceptance rules. $^\dagger$ DA3-estimated scale; others source-verified.
  Scales are fixed across methods and all scenes enter the means.
  Development results; bold marks the best comparable rate and
  displayed ties. No-result rates are descriptive.}
  \label{tab:nine_scene_detail}
  \fontsize{8}{9.5}\selectfont
  \setlength{\tabcolsep}{2pt}
  \renewcommand{\arraystretch}{1.02}
  \textbf{(a) Text-conditioned view retrieval}\par\smallskip
  \begin{tabular*}{\textwidth}{@{\extracolsep{\fill}}p{75pt}p{72pt}*{10}{c}@{}}
    \toprule
    Method & Metric (\%) & \shortstack{Block\\Small} & \shortstack{Block\\A} & NYC & SF & Citysample & Colosseum & Elvenruin & Park & Road & \shortstack{Scene\\mean} \\
    \midrule
    Queries & $n$ & 20 & 20 & 20 & 20 & 20 & 20 & 20 & 20 & 20 & --- \\
    \midrule
    \multirow{3}{*}{\shortstack[l]{CityLoC~\cite{ma2025cityloc}\\without refinement}} & Top-1 success $\uparrow$ & 21.7 & 15.0 & 21.7 & 43.3 & 30.0 & 33.3 & 20.0 & 20.0 & 20.0 & 25.0 \\
     & Top-5 success $\uparrow$ & 25.0 & 21.7 & 21.7 & 45.0 & 33.3 & 40.0 & 23.3 & 35.0 & 31.7 & 30.7 \\
     & No result & 0.0 & 0.0 & 0.0 & 0.0 & 0.0 & 0.0 & 0.0 & 0.0 & 0.0 & 0.0 \\
    \midrule
    \multirow{3}{*}{\shortstack[l]{CityLoC~\cite{ma2025cityloc}\\with refinement}} & Top-1 success $\uparrow$ & 21.7 & 15.0 & 16.7 & 48.3 & 35.0 & 38.3 & 20.0 & 16.7 & 23.3 & 26.1 \\
     & Top-5 success $\uparrow$ & 25.0 & 18.3 & 30.0 & 56.7 & 43.3 & 50.0 & 25.0 & 31.7 & 31.7 & 34.6 \\
     & No result & 26.7 & 35.0 & 15.0 & 1.7 & 41.7 & 8.3 & 40.0 & 8.3 & 1.7 & 19.8 \\
    \midrule
    \multirow{3}{*}{\textbf{QGVI-T}} & Top-1 success $\uparrow$ & \textbf{65.0} & \textbf{60.0} & \textbf{60.0} & \textbf{55.0} & \textbf{50.0} & \textbf{50.0} & \textbf{80.0} & \textbf{40.0} & \textbf{65.0} & \textbf{58.3} \\
     & Top-5 success $\uparrow$ & \textbf{90.0} & \textbf{80.0} & \textbf{85.0} & \textbf{85.0} & \textbf{65.0} & \textbf{80.0} & \textbf{90.0} & \textbf{55.0} & \textbf{75.0} & \textbf{78.3} \\
     & No result & 0.0 & 0.0 & 0.0 & 0.0 & 0.0 & 0.0 & 0.0 & 0.0 & 0.0 & 0.0 \\
    \bottomrule
  \end{tabular*}
  \par\medskip
  \textbf{(b) RGB image-query camera localization}\par\smallskip
  \begin{tabular*}{\textwidth}{@{\extracolsep{\fill}}p{75pt}p{72pt}*{10}{c}@{}}
    \toprule
    Method & Metric (\%) & \shortstack{Block\\Small} & \shortstack{Block\\A} & NYC & SF & Citysample & Colosseum & Elvenruin & Park & Road & \shortstack{Scene\\mean} \\
    \midrule
    In-map queries & $n$ & 139 & 602 & 180 & 174 & 142 & 126 & 118 & 228 & 130 & --- \\
    Out-of-map queries & $n$ & 139 & 602 & 180 & 174 & 142 & 126 & 118 & 228 & 130 & --- \\
    Scale & m / native unit & 25 & 25 & 1 & 1 & 24.777$^\dagger$ & 12.287$^\dagger$ & 14.135$^\dagger$ & 1.040$^\dagger$ & 1.339$^\dagger$ & --- \\
    \midrule
    \multirow{4}{*}{LoGS~\cite{cheng2025logs}} & Pose 1\,m/2$^\circ$ $\uparrow$ & \textbf{90.6} & \textbf{93.4} & \textbf{100.0} & \textbf{100.0} & \textbf{100.0} & \textbf{91.3} & \textbf{100.0} & \textbf{100.0} & \textbf{100.0} & \textbf{97.3} \\
 & Pose 5\,m/5$^\circ$ $\uparrow$ & \textbf{92.1} & \textbf{93.5} & \textbf{100.0} & \textbf{100.0} & \textbf{100.0} & \textbf{91.3} & \textbf{100.0} & \textbf{100.0} & \textbf{100.0} & \textbf{97.4} \\
 & False accept $\downarrow$ & 100.0 & 100.0 & 100.0 & 100.0 & 100.0 & 100.0 & 100.0 & 100.0 & 100.0 & 100.0 \\
 & False reject $\downarrow$ & \textbf{0.0} & \textbf{0.0} & \textbf{0.0} & \textbf{0.0} & \textbf{0.0} & \textbf{0.0} & \textbf{0.0} & \textbf{0.0} & \textbf{0.0} & \textbf{0.0} \\
    \midrule
    \multirow{4}{*}{SplatHLoc~\cite{tao2026splathloc}} & Pose 1\,m/2$^\circ$ $\uparrow$ & 65.5 & 12.6 & \textbf{100.0} & \textbf{100.0} & 62.7 & 82.5 & 91.5 & 84.6 & 84.6 & 76.0 \\
 & Pose 5\,m/5$^\circ$ $\uparrow$ & 69.1 & 18.1 & \textbf{100.0} & \textbf{100.0} & 78.9 & 83.3 & 94.9 & 88.6 & 87.7 & 80.1 \\
 & False accept $\downarrow$ & 100.0 & 98.0 & 100.0 & 100.0 & 100.0 & 100.0 & 100.0 & 100.0 & 100.0 & 99.8 \\
 & False reject $\downarrow$ & \textbf{0.0} & 0.8 & \textbf{0.0} & \textbf{0.0} & \textbf{0.0} & \textbf{0.0} & \textbf{0.0} & \textbf{0.0} & \textbf{0.0} & 0.1 \\
    \midrule
    \multirow{4}{*}{\textbf{QGVI-I}} & Pose 1\,m/2$^\circ$ $\uparrow$ & \textbf{90.6} & \textbf{93.4} & \textbf{100.0} & \textbf{100.0} & \textbf{100.0} & \textbf{91.3} & \textbf{100.0} & \textbf{100.0} & \textbf{100.0} & \textbf{97.3} \\
 & Pose 5\,m/5$^\circ$ $\uparrow$ & \textbf{92.1} & \textbf{93.5} & \textbf{100.0} & \textbf{100.0} & \textbf{100.0} & \textbf{91.3} & \textbf{100.0} & \textbf{100.0} & \textbf{100.0} & \textbf{97.4} \\
 & False accept $\downarrow$ & \textbf{0.000} & \textbf{0.332} & \textbf{1.111} & \textbf{4.023} & \textbf{1.408} & \textbf{0.000} & \textbf{0.000} & \textbf{0.000} & \textbf{0.000} & \textbf{0.764} \\
 & False reject $\downarrow$ & 3.597 & 2.492 & \textbf{0.000} & \textbf{0.000} & \textbf{0.000} & 0.794 & \textbf{0.000} & \textbf{0.000} & \textbf{0.000} & 0.765 \\
    \bottomrule
  \end{tabular*}
\end{table*}

\section{Optional Text Verification}
\label{sec:supp_text_verification}
Historical QGVI-T + verification evaluates the 20 fusion-shortlisted views using a VLM.
The verifier receives the rendered image and query, excluding the caption.
It enumerates mandatory constraints, including conjunctions, counts,
attributes, and spatial relations, and assigns each a label with supporting
visual evidence: supported, contradicted, or unclear. A view qualifies only
if the verifier reports a match and every reported requirement is supported.
Response validation overrides a match verdict when any requirement is
contradicted or unclear. Contradiction produces a non-match decision;
insufficient evidence produces an uncertain decision.

Eligible views are ranked by verifier relevance, with shortlist order
breaking ties. Scores are not calibrated probabilities. We return at most
five distinct places with unchanged stored poses, or no result when none
qualifies. This abstention concerns the shortlist and does not establish
absence throughout the scene. Query processing uses cached renders without
new rendering calls.

\section{Matched Text Retrieval Controls}
\label{sec:matched_text_supp}
CLIP, BM25, MiniLM, and historical QGVI-T fusion use the same 30,522 rendered views,
180 descriptions, and place selection: 64 candidates, a 20-view shortlist,
and five distinct returned places. The three caption methods share the same
captions; all four omit verification. Each reviewer judges identical query--image
pairs once, using the rubric in Sec.~\ref{sec:supp_human}.
\begin{table*}[t]
\centering
\caption{\textbf{Matched text retrieval by scene.}
(a) GPT joint relevance-and-validity Top-1/Top-5 success bounds (\%),
with 20 queries per scene and 180 pooled. GPT completed all 2,825 distinct
query--image assessments. (b) Separate pooled Kimi assessment: 2,660/2,825
judgments completed; 165 missing judgments remain unresolved.
All methods return five places for every query. Intervals include uncertain
labels and, for Kimi, missing judgments; they are not confidence intervals.
Bold marks the highest conservative lower endpoint per metric and row in
(a), and per metric in (b), including ties. Overlap limits ranking claims.}
\label{tab:matched_text_scene}
\footnotesize
\setlength{\tabcolsep}{3pt}
\textit{(a) GPT assessment}\par\smallskip
\begin{tabular}{@{}l*{8}{c}@{}}\toprule
Scene & \multicolumn{2}{c}{CLIP} & \multicolumn{2}{c}{BM25} & \multicolumn{2}{c}{MiniLM} & \multicolumn{2}{c}{QGVI-T (fusion)} \\
\cmidrule(lr){2-3}\cmidrule(lr){4-5}\cmidrule(lr){6-7}\cmidrule(lr){8-9}
 & Top-1 $\uparrow$ & Top-5 $\uparrow$ & Top-1 $\uparrow$ & Top-5 $\uparrow$ & Top-1 $\uparrow$ & Top-5 $\uparrow$ & Top-1 $\uparrow$ & Top-5 $\uparrow$ \\\midrule
Block Small & $35.0\text{--}40.0$ & $55.0\text{--}70.0$ & $60.0\text{--}75.0$ & $70.0\text{--}95.0$ & $60.0\text{--}70.0$ & $\mathbf{80.0}\text{--}90.0$ & $\mathbf{65.0}\text{--}70.0$ & $\mathbf{80.0}\text{--}90.0$ \\
Block A & $30.0\text{--}60.0$ & $45.0\text{--}70.0$ & $\mathbf{60.0}\text{--}85.0$ & $\mathbf{70.0}\text{--}95.0$ & $55.0\text{--}65.0$ & $60.0\text{--}80.0$ & $50.0\text{--}60.0$ & $65.0\text{--}85.0$ \\
NYC & $40.0\text{--}55.0$ & $75.0\text{--}100.0$ & $\mathbf{65.0}\text{--}80.0$ & $\mathbf{80.0}\text{--}95.0$ & $55.0\text{--}75.0$ & $75.0\text{--}95.0$ & $\mathbf{65.0}\text{--}85.0$ & $70.0\text{--}95.0$ \\
SF & $60.0\text{--}80.0$ & $85.0\text{--}100.0$ & $65.0\text{--}80.0$ & $90.0\text{--}95.0$ & $\mathbf{80.0}\text{--}90.0$ & $90.0\text{--}95.0$ & $70.0\text{--}90.0$ & $\mathbf{95.0}\text{--}100.0$ \\
Citysample & $45.0\text{--}65.0$ & $60.0\text{--}80.0$ & $45.0\text{--}75.0$ & $\mathbf{75.0}\text{--}85.0$ & $50.0\text{--}75.0$ & $50.0\text{--}80.0$ & $\mathbf{55.0}\text{--}80.0$ & $65.0\text{--}90.0$ \\
Colosseum & $40.0\text{--}50.0$ & $55.0\text{--}85.0$ & $\mathbf{75.0}\text{--}85.0$ & $\mathbf{85.0}\text{--}90.0$ & $55.0\text{--}75.0$ & $75.0\text{--}95.0$ & $60.0\text{--}80.0$ & $80.0\text{--}90.0$ \\
Elvenruin & $50.0\text{--}70.0$ & $85.0\text{--}90.0$ & $\mathbf{85.0}\text{--}90.0$ & $\mathbf{95.0}\text{--}100.0$ & $75.0\text{--}85.0$ & $\mathbf{95.0}\text{--}100.0$ & $\mathbf{85.0}\text{--}95.0$ & $\mathbf{95.0}\text{--}100.0$ \\
Park & $30.0\text{--}40.0$ & $45.0\text{--}55.0$ & $\mathbf{45.0}\text{--}60.0$ & $\mathbf{50.0}\text{--}75.0$ & $40.0\text{--}65.0$ & $\mathbf{50.0}\text{--}80.0$ & $35.0\text{--}55.0$ & $40.0\text{--}75.0$ \\
Road & $25.0\text{--}30.0$ & $55.0\text{--}75.0$ & $\mathbf{65.0}\text{--}80.0$ & $\mathbf{80.0}\text{--}100.0$ & $40.0\text{--}65.0$ & $60.0\text{--}95.0$ & $\mathbf{65.0}\text{--}85.0$ & $75.0\text{--}100.0$ \\
\midrule
Pooled & $39.4\text{--}54.4$ & $62.2\text{--}80.6$ & $\mathbf{62.8}\text{--}78.9$ & $\mathbf{77.2}\text{--}92.2$ & $56.7\text{--}73.9$ & $70.6\text{--}90.0$ & $61.1\text{--}77.8$ & $73.9\text{--}91.7$ \\
\bottomrule\end{tabular}
\par\medskip
\textit{(b) Kimi assessment, pooled over the same 180 queries}\par\smallskip
\begin{tabular}{@{}lccc@{}}\toprule
Method & Top-1 (\%) $\uparrow$ & Top-5 (\%) $\uparrow$ & Net utility $100U_1$ $\uparrow$ \\\midrule
CLIP & $32.2\text{--}57.2$ & $57.8\text{--}90.6$ & $[-35.6,\,14.4]$ \\
BM25 & $\mathbf{55.6}\text{--}75.6$ & $\mathbf{75.6}\text{--}96.1$ & $[\mathbf{11.1},\,51.1]$ \\
MiniLM & $50.6\text{--}76.7$ & $66.1\text{--}93.9$ & $[1.1,\,53.3]$ \\
QGVI-T (fusion) & $52.2\text{--}73.9$ & $72.8\text{--}93.3$ & $[4.4,\,47.8]$ \\
\bottomrule\end{tabular}
\end{table*}

\section{View-Bank Composition and Novel Views}
\label{sec:view_bank_supp}

We vary the candidate observations while retaining the same 180 development
descriptions, historical fusion scorer, and five-place output budget.
The original-image bank contains 18,570 training images, the grid bank
contains 30,522 Gaussian renders, and their union contains 49,092 images.
All returned cameras are assessed using their Gaussian renderings,
including cameras retrieved through original-image captions.
Thus the comparison measures semantic access to the supplied asset.

GPT completed all 3,263 distinct query--image judgments for the 64 bank
conditions and all 1,329 judgments for the 18 original/union conditions.
The following tables use this completed panel. Kimi returned 3,023/3,263
and 1,200/1,329 valid judgments, respectively; the remaining 240 and 129
pairs exhausted the retry budget. These incomplete panels are not used
to assert a second-reviewer confirmation of the bank comparisons.
Uncertain labels remain unresolved even in the completed GPT panel.

\begin{table*}[t]
\centering
\caption{\textbf{Candidate-bank comparison by scene (GPT).} Each scene has 20 queries; pooled rows use all 180. Entries are lower--upper success bounds in percent, not confidence intervals. Top-1 requires a successful first result; Top-5 requires at least one successful result among five. Bold marks the highest confirmed-success lower bound in each scene and metric, including ties; overlapping bounds limit ranking claims. The original bank uses training images, and the union retains those images while adding the grid.}
\label{tab:view_bank_scenes}
\begingroup\small\setlength{\tabcolsep}{5pt}
\begin{tabular}{@{}lrrrrrr@{}}\toprule
Scene & \multicolumn{2}{c}{Original images} & \multicolumn{2}{c}{Grid renders} & \multicolumn{2}{c}{Original + grid} \\
\cmidrule(lr){2-3}\cmidrule(lr){4-5}\cmidrule(lr){6-7}
 & Top-1 & Top-5 & Top-1 & Top-5 & Top-1 & Top-5 \\\midrule
Block Small & $50.0\text{--}50.0$ & $75.0\text{--}85.0$ & $\mathbf{65.0}\text{--}70.0$ & $\mathbf{80.0}\text{--}90.0$ & $55.0\text{--}55.0$ & $75.0\text{--}80.0$ \\
Block A & $40.0\text{--}65.0$ & $\mathbf{70.0}\text{--}85.0$ & $\mathbf{50.0}\text{--}60.0$ & $65.0\text{--}85.0$ & $45.0\text{--}70.0$ & $\mathbf{70.0}\text{--}90.0$ \\
NYC & $\mathbf{65.0}\text{--}95.0$ & $\mathbf{75.0}\text{--}95.0$ & $\mathbf{65.0}\text{--}85.0$ & $70.0\text{--}95.0$ & $\mathbf{65.0}\text{--}90.0$ & $\mathbf{75.0}\text{--}95.0$ \\
SF & $75.0\text{--}85.0$ & $85.0\text{--}100.0$ & $70.0\text{--}90.0$ & $\mathbf{95.0}\text{--}100.0$ & $\mathbf{80.0}\text{--}90.0$ & $85.0\text{--}100.0$ \\
City Sample & $\mathbf{55.0}\text{--}80.0$ & $\mathbf{65.0}\text{--}90.0$ & $\mathbf{55.0}\text{--}80.0$ & $\mathbf{65.0}\text{--}90.0$ & $\mathbf{55.0}\text{--}80.0$ & $\mathbf{65.0}\text{--}95.0$ \\
Colosseum & $\mathbf{65.0}\text{--}80.0$ & $\mathbf{85.0}\text{--}90.0$ & $60.0\text{--}80.0$ & $80.0\text{--}90.0$ & $\mathbf{65.0}\text{--}80.0$ & $80.0\text{--}95.0$ \\
Elven Ruin & $70.0\text{--}75.0$ & $80.0\text{--}95.0$ & $\mathbf{85.0}\text{--}95.0$ & $\mathbf{95.0}\text{--}100.0$ & $80.0\text{--}85.0$ & $\mathbf{95.0}\text{--}95.0$ \\
Park & $\mathbf{45.0}\text{--}60.0$ & $\mathbf{55.0}\text{--}85.0$ & $35.0\text{--}55.0$ & $40.0\text{--}75.0$ & $\mathbf{45.0}\text{--}65.0$ & $\mathbf{55.0}\text{--}85.0$ \\
Road & $55.0\text{--}70.0$ & $70.0\text{--}90.0$ & $\mathbf{65.0}\text{--}85.0$ & $\mathbf{75.0}\text{--}100.0$ & $50.0\text{--}75.0$ & $70.0\text{--}90.0$ \\
\midrule
Pooled & $57.8\text{--}73.3$ & $73.3\text{--}90.6$ & $\mathbf{61.1}\text{--}77.8$ & $73.9\text{--}91.7$ & $60.0\text{--}76.7$ & $\mathbf{74.4}\text{--}91.7$ \\
\bottomrule\end{tabular}\endgroup
\end{table*}

On all nine scenes, adding the grid changes confirmed Top-5 success
from 132/180 to 134/180. Its paired change lies in $[-10.0,+12.2]$
percentage points under unresolved labels. This result does not establish
a consistent improvement from novel-view augmentation.

\subsection{Appearance and Candidate Coverage}
The controlled study uses Block A and SF. A same-pose bank replaces each
training image by a Gaussian rendering at the identical camera.
The expanded-render bank adds the grid to these training-camera renders.
It differs from the nine-scene original-plus-grid bank, which retains
the original training images. Table~\ref{tab:view_bank_two_scene} reports
both the appearance change and the subsequent coverage change.

\begin{table*}[t]
\centering
\caption{\textbf{Same-pose appearance and expanded-render controls (GPT).} Block A and SF have $N=20$ queries each; pooled rows have $N=40$. Same-pose uses Gaussian renders at all training cameras; expanded-render adds grid renders. Bounds are label-uncertainty intervals. Changes are paired Top-5 differences in percentage points against the original-image bank (O) and same-pose bank (R). Bold compares the two rendered-bank conditions within each cohort.}
\label{tab:view_bank_two_scene}
\begingroup\small\setlength{\tabcolsep}{5pt}
\begin{tabular}{@{}llrrrrr@{}}\toprule
Cohort & Candidate bank & Images & Top-1 (\%) & Top-5 (\%) & $\Delta$ Top-5 vs O & $\Delta$ Top-5 vs R \\\midrule
Block A & Same-pose renders & 5,138 & $50.0\text{--}65.0$ & $65.0\text{--}85.0$ & $[-20.0,\,+15.0]$ & $[+0.0,\,+0.0]$ \\
Block A & Expanded renders & 18,232 & $\mathbf{55.0}\text{--}75.0$ & $\mathbf{70.0}\text{--}90.0$ & $[-15.0,\,+20.0]$ & $[-5.0,\,+15.0]$ \\
\midrule
SF & Same-pose renders & 666 & $\mathbf{75.0}\text{--}85.0$ & $\mathbf{90.0}\text{--}95.0$ & $[-10.0,\,+10.0]$ & $[+0.0,\,+0.0]$ \\
SF & Expanded renders & 3,549 & $\mathbf{75.0}\text{--}85.0$ & $\mathbf{90.0}\text{--}95.0$ & $[-10.0,\,+10.0]$ & $[-5.0,\,+5.0]$ \\
\midrule
Pooled & Same-pose renders & 5,804 & $62.5\text{--}75.0$ & $77.5\text{--}90.0$ & $[-15.0,\,+12.5]$ & $[+0.0,\,+0.0]$ \\
Pooled & Expanded renders & 21,781 & $\mathbf{65.0}\text{--}80.0$ & $\mathbf{80.0}\text{--}92.5$ & $[-12.5,\,+15.0]$ & $[-5.0,\,+10.0]$ \\
\bottomrule\end{tabular}\endgroup
\end{table*}

\subsection{Replacement and Density Controls}
Replacement substitutes a fraction of the training-camera renders with
grid renders while retaining the original image count and domain counts.
Growth retains every training-camera render and adds a fraction of the
eligible grid. Both use three fixed geometry-based sampling seeds.
NYC and SF have no aerial grid; their original aerial observations remain
in the nine-scene comparison. Within the two-scene replacement study,
this constraint limits SF to actual replacement fractions of 11.9\%,
23.7\%, and 35.4\% at nominal settings of 25\%, 50\%, and 75\%.

\begin{table*}[t]
\centering
\caption{\textbf{All replacement and growth settings (GPT).} Every row averages all three fixed seeds on the same 20 queries; seeds do not increase the query denominator. Success columns average the lower and upper label-uncertainty endpoints. The seed range is the minimum--maximum confirmed Top-5 rate. $\Delta$ gives mean paired Top-5 uncertainty endpoints against the same-pose bank, in percentage points. For replacement, actual fraction is the proportion of training entries replaced; for growth, it is the proportion of eligible grid entries added. Bold marks the largest mean confirmed rate within each scene and policy.}
\label{tab:view_bank_density}
\begingroup\footnotesize\setlength{\tabcolsep}{3.1pt}
\begin{tabular}{@{}llrrrrrrr@{}}\toprule
Scene & Policy & \shortstack{Nominal\\(\%)} & \shortstack{Actual\\(\%)} & Images & \shortstack{Mean Top-1\\bounds (\%)} & \shortstack{Mean Top-5\\bounds (\%)} & \shortstack{Top-5 seed\\range (\%)} & $\Delta$ Top-5 \\\midrule
Block A & Replace & 0 & 0.0 & 5,138 & $50.0\text{--}65.0$ & $65.0\text{--}85.0$ & $65.0\text{--}65.0$ & $[+0.0,\,+0.0]$ \\
Block A & Replace & 25 & 25.0 & 5,138 & $55.0\text{--}65.0$ & $68.3\text{--}85.0$ & $65.0\text{--}70.0$ & $[-13.3,\,+13.3]$ \\
Block A & Replace & 50 & 50.0 & 5,138 & $\mathbf{56.7}\text{--}70.0$ & $68.3\text{--}88.3$ & $65.0\text{--}75.0$ & $[-15.0,\,+20.0]$ \\
Block A & Replace & 75 & 75.0 & 5,138 & $48.3\text{--}61.7$ & $\mathbf{71.7}\text{--}91.7$ & $70.0\text{--}75.0$ & $[-13.3,\,+26.7]$ \\
\midrule
Block A & Grow & 25 & 25.0 & 8,412 & $50.0\text{--}66.7$ & $68.3\text{--}86.7$ & $65.0\text{--}70.0$ & $[-5.0,\,+10.0]$ \\
Block A & Grow & 50 & 50.0 & 11,685 & $51.7\text{--}68.3$ & $68.3\text{--}86.7$ & $65.0\text{--}70.0$ & $[-6.7,\,+10.0]$ \\
Block A & Grow & 100 & 100.0 & 18,232 & $\mathbf{55.0}\text{--}75.0$ & $\mathbf{70.0}\text{--}90.0$ & $70.0\text{--}70.0$ & $[-5.0,\,+15.0]$ \\
\midrule
SF & Replace & 0 & 0.0 & 666 & $75.0\text{--}85.0$ & $90.0\text{--}95.0$ & $90.0\text{--}90.0$ & $[+0.0,\,+0.0]$ \\
SF & Replace & 25 & 11.9 & 666 & $78.3\text{--}85.0$ & $90.0\text{--}95.0$ & $90.0\text{--}90.0$ & $[+0.0,\,+0.0]$ \\
SF & Replace & 50 & 23.7 & 666 & $\mathbf{83.3}\text{--}88.3$ & $90.0\text{--}95.0$ & $90.0\text{--}90.0$ & $[+0.0,\,+0.0]$ \\
SF & Replace & 75 & 35.4 & 666 & $76.7\text{--}85.0$ & $\mathbf{91.7}\text{--}98.3$ & $90.0\text{--}95.0$ & $[+0.0,\,+5.0]$ \\
\midrule
SF & Grow & 25 & 25.0 & 1,387 & $71.7\text{--}83.3$ & $\mathbf{90.0}\text{--}95.0$ & $90.0\text{--}90.0$ & $[-3.3,\,+3.3]$ \\
SF & Grow & 50 & 50.0 & 2,108 & $71.7\text{--}83.3$ & $\mathbf{90.0}\text{--}95.0$ & $90.0\text{--}90.0$ & $[-5.0,\,+5.0]$ \\
SF & Grow & 100 & 100.0 & 3,549 & $\mathbf{75.0}\text{--}85.0$ & $\mathbf{90.0}\text{--}95.0$ & $90.0\text{--}90.0$ & $[-5.0,\,+5.0]$ \\
\bottomrule\end{tabular}\endgroup
\end{table*}

All paired Top-5 intervals in these controls include zero. The variation
across fractions and seeds does not support a monotonic retrieval benefit
from a denser bank. These observations concern the sampled banks and
the exposed development descriptions.

\subsection{Paired Results by Description Source}
The descriptions were authored from 74 held-out original dataset images
and 106 novel Gaussian renders. Original dataset images include both real
and synthetic source datasets. The source groups have different scene
mixtures: Park contributes 18 original-source and two render-source
queries, whereas Colosseum contributes one and 19. The source-stratified
differences therefore do not isolate a causal effect of query origin.

\begin{table}[t]
\centering
\caption{\textbf{Paired original-plus-grid versus original-only outcomes (GPT).} W/L/S/U denote confirmed wins, confirmed losses, unchanged outcomes, and unresolved changes. Identical unresolved answer sets have zero paired change and enter S. Bounds are possible paired changes under unresolved labels, not confidence intervals.}
\label{tab:view_bank_source_pairs}
\begingroup\footnotesize\setlength{\tabcolsep}{3pt}
\begin{tabular}{@{}llrr@{}}\toprule
Source ($N$) & Metric & W/L/S/U & $\Delta$ (pp) \\\midrule
Original (74) & Top-1 & 1/1/67/5 & $[-4.1,\,+5.4]$ \\
Original (74) & Top-5 & 0/0/64/10 & $[-9.5,\,+5.4]$ \\
Render (106) & Top-1 & 4/0/89/13 & $[-3.8,\,+12.3]$ \\
Render (106) & Top-5 & 1/1/82/22 & $[-10.4,\,+17.0]$ \\
\bottomrule\end{tabular}\endgroup
\end{table}

Confirmed Top-5 successes change from 53 to 52 in the original-source
group and from 79 to 82 in the render-source group. Changes in confirmed
counts can involve uncertain outcomes, so they need not equal W minus L.
Both source groups retain paired uncertainty intervals containing zero.

\section{Matched rejection controls and photometric refinement}
\label{sec:calibrated_rgb_supp}
We apply alternative acceptance rules to the same saved poses and calibrate
each rule using only the 490 Block A/SF validation queries (245 in-map and
245 out-of-map). Count thresholds are $\{0,4,8,12,20,40,80\}$; ratio thresholds
are $\{0,.05,.1,.2,.35,.5,.7\}$. A feasible rule must retain every pose in
each validation-correct set at $1\,\mathrm{m}/2^\circ$ and
$5\,\mathrm{m}/5^\circ$.
Among feasible rules we maximize out-of-map rejection, then in-map acceptance,
and break remaining ties in favor of fewer active gates and a lower threshold.
We freeze these rules before scoring the 3,678 queries across nine scenes;
the cohort has been used in earlier development and is not a new independent
test set. Validation retention does not guarantee retention on these queries.

\begin{table*}[t]
\centering
\caption{Calibrated RGB rejection controls. All entries pool the nine scenes
(1,839 in-map and 1,839 out-of-map queries). FAR and FRR use the out-of-map and
in-map denominators, respectively; pose success includes rejected in-map queries
as failures. Precision is correct $1\,\mathrm{m}/2^\circ$ poses divided by all
accepted queries, including out-of-map returns. $100U_1=100(S-W)/3678$.
Best values are bold within each LoGS pose family, with ties included;
boldface does not imply joint optimality across columns. The SplatHLoc row has
its own native poses. $\dagger$ The added threshold is zero because native
eligibility already requires four final inliers; thresholds zero and four are
equivalent. All gates preserve retained pose coordinates.}
\label{tab:all_calibrated_rgb}
\setlength{\tabcolsep}{4pt}
\small
\begin{tabular}{lrrrrrrr}
\toprule
Acceptance rule & Threshold & FAR$\downarrow$ & FRR$\downarrow$ &
\multicolumn{2}{c}{Pose success ($\%$)$\uparrow$} & Precision$\uparrow$ & $100U_1\uparrow$ \\
 & & ($\%$) & ($\%$) & $1\,\mathrm{m}/2^\circ$ & $5\,\mathrm{m}/5^\circ$ & ($\%$) & \\
\midrule
\multicolumn{8}{l}{\textit{Fixed LoGS initializer poses (no photometric refinement)}} \\
Return every pose & --- & 100.000 & \textbf{0.000} & \textbf{96.52} & \textbf{96.68} & 48.26 & -3.48 \\
PnP success only & --- & 99.674 & 0.054 & \textbf{96.52} & \textbf{96.68} & 48.35 & -3.29 \\
Raw inlier count & 40 & 1.305 & 1.033 & \textbf{96.52} & \textbf{96.68} & 96.26 & 46.38 \\
Raw match inlier ratio & 0.1 & 79.445 & 0.761 & 96.36 & 96.47 & 53.93 & 7.01 \\
Pair inlier ratio & 0.1 & 85.101 & 0.435 & \textbf{96.52} & \textbf{96.68} & 52.27 & 4.19 \\
Unique-keypoint inlier ratio & 0.2 & 24.035 & 0.979 & 96.41 & 96.57 & 78.35 & 34.88 \\
Unique inlier count (refit) & 40 & \textbf{0.707} & 1.142 & \textbf{96.52} & \textbf{96.68} & \textbf{96.94} & \textbf{46.74} \\
QGVI-I & 40 & \textbf{0.707} & 1.142 & \textbf{96.52} & \textbf{96.68} & \textbf{96.94} & \textbf{46.74} \\
\midrule
\multicolumn{8}{l}{\textit{Native LoGS refined poses (up to 20 photometric steps)}} \\
Initial raw inlier count & 40 & 1.305 & 1.033 & \textbf{96.52} & \textbf{96.68} & 96.26 & 46.38 \\
Initial raw match inlier ratio & 0.1 & 79.445 & 0.761 & 96.36 & 96.47 & 53.93 & 7.01 \\
Final pair inlier count & 40 & 1.251 & 1.142 & \textbf{96.52} & \textbf{96.68} & 96.41 & 46.47 \\
Final pair inlier ratio & 0.1 & 81.185 & \textbf{0.435} & \textbf{96.52} & \textbf{96.68} & 53.40 & 6.14 \\
QGVI-I + photo. & 40 & \textbf{0.544} & 1.142 & \textbf{96.52} & \textbf{96.68} & \textbf{97.10} & \textbf{46.82} \\
Final unique-keypoint ratio & 0.2 & 22.023 & 0.979 & 96.41 & 96.57 & 79.65 & 35.89 \\
\midrule
\multicolumn{8}{l}{\textit{Native SplatHLoc final poses}} \\
Final inlier count & $0^{\dagger}$ & 99.347 & 0.272 & 61.17 & 65.47 & 30.73 & -38.36 \\
\bottomrule
\end{tabular}
\end{table*}

\begin{table*}[t]
\centering
\caption{Scene-level comparison of the calibrated distinct-keypoint gates.
G denotes the geometric QGVI-I default; R denotes QGVI-I + photo. Both
require at least 40 distinct inlier query keypoints at their returned pose
and have identical pose-success and FRR entries in each scene. Precision
and utility use the strict pose threshold. Bold highlights the better
G/R value within each scene, including ties. Estimated DA3 scales
for five scenes qualify the metre-based pose thresholds.}
\label{tab:calibrated_rgb_scene_detail}
\small
\setlength{\tabcolsep}{4pt}
\begin{tabular}{lrrrrrrrrrr}
\toprule
Scene & $N_{\rm in}$ & \multicolumn{2}{c}{Pose success ($\%$)$\uparrow$} & FRR ($\%$)$\downarrow$ &
\multicolumn{2}{c}{FAR ($\%$)$\downarrow$} & \multicolumn{2}{c}{Precision ($\%$)$\uparrow$} & \multicolumn{2}{c}{$100U_1\uparrow$} \\
 & & $1\,\mathrm{m}/2^\circ$ & $5\,\mathrm{m}/5^\circ$ & & G & R & G & R & G & R \\
\midrule
Block Small & 139 & 90.65 & 92.09 & 3.597 & \textbf{0.000} & \textbf{0.000} & \textbf{94.03} & \textbf{94.03} & \textbf{42.45} & \textbf{42.45} \\
Block A & 602 & 93.36 & 93.52 & 2.492 & \textbf{0.332} & \textbf{0.332} & \textbf{95.42} & \textbf{95.42} & \textbf{44.44} & \textbf{44.44} \\
NYC & 180 & 100.00 & 100.00 & 0.000 & 1.111 & \textbf{0.556} & 98.90 & \textbf{99.45} & 49.44 & \textbf{49.72} \\
SF & 174 & 100.00 & 100.00 & 0.000 & 4.023 & \textbf{2.874} & 96.13 & \textbf{97.21} & 47.99 & \textbf{48.56} \\
City Sample & 142 & 100.00 & 100.00 & 0.000 & \textbf{1.408} & \textbf{1.408} & \textbf{98.61} & \textbf{98.61} & \textbf{49.30} & \textbf{49.30} \\
Colosseum & 126 & 91.27 & 91.27 & 0.794 & \textbf{0.000} & \textbf{0.000} & \textbf{92.00} & \textbf{92.00} & \textbf{41.67} & \textbf{41.67} \\
Elven Ruin & 118 & 100.00 & 100.00 & 0.000 & \textbf{0.000} & \textbf{0.000} & \textbf{100.00} & \textbf{100.00} & \textbf{50.00} & \textbf{50.00} \\
Park & 228 & 100.00 & 100.00 & 0.000 & \textbf{0.000} & \textbf{0.000} & \textbf{100.00} & \textbf{100.00} & \textbf{50.00} & \textbf{50.00} \\
Road & 130 & 100.00 & 100.00 & 0.000 & \textbf{0.000} & \textbf{0.000} & \textbf{100.00} & \textbf{100.00} & \textbf{50.00} & \textbf{50.00} \\
\bottomrule
\end{tabular}
\end{table*}

The raw match ratio divides RANSAC inliers by all saved raw matches; the pair
ratio instead uses the deduplicated query-keypoint/landmark pairs supplied to
PnP. A unique-keypoint ratio uses distinct inlier query keypoints over distinct
matched query keypoints. For refined LoGS, final support is recomputed at the
unchanged refined pose using positive depth and a $12$-pixel reprojection bound.
Initializer-score gates in that panel select already-refined poses and do not
save refinement computation. SplatHLoc does not save a correspondence
denominator, so we evaluate its observed final inlier count only.

On fixed initializer poses, recalibration selects a distinct count of 40. Relative to
a raw inlier-count threshold of 40 on the same initializer, it removes 11
additional out-of-map returns and two incorrect in-map returns, without losing
a correct pose. This geometric route is the QGVI-I default. The optional
QGVI-I + photo. ablation applies the final distinct-keypoint gate to full
LoGS and gives 10 false accepts instead of 13, at the same 21 in-map
rejections and 1,775 strict successes. It uses native LoGS refinement,
without a new pose optimizer. These comparisons use the exposed development
cohort rather than an independent final test. For SplatHLoc, a threshold of eight would reject more negatives
but lose one validation pose correct at the looser threshold; the no-loss
criterion therefore selects its native acceptance point. This is a constraint
on the chosen operating point, not evidence that SplatHLoc cannot reject.

Threshold-swept risk--coverage summaries qualify the fixed operating-point
comparison. On the identical 3,671 PnP-eligible initializer poses, the grouped
right-step area is 0.17638 for unique inlier count, 0.17492 for raw inlier
count, and 0.16641 for the unique-keypoint ratio. The unique count therefore
has the best reported validation-constrained operating point, but not the
lowest integrated risk. These areas cover the observed eligible range only
($0$--$99.81\%$ coverage), without breaking score ties; they are not compared
across methods with different poses or eligibility.

\paragraph{Paired photometric diagnostic.}
We additionally compare the saved initializer with native LoGS
refinement (up to 20 steps) for every RGB query. All 3,678 pose matrices change. Among the
1,839 in-map queries, strict success remains 1,775 and loose success remains
1,778, with zero individual gains or losses at either threshold. Continuous
pose errors nevertheless change: median translation error increases from
0.01191 to 0.02757\,m,
and median rotation error from $0.02329^\circ$ to
$0.08531^\circ$. Translation error improves for 405 queries and worsens for
1,434; rotation error improves for 129 and worsens for 1,710
(a $10^{-9}$ tolerance in metres or degrees). These aggregate observations
on the existing development cohort show no threshold-success gain from
refinement and worse median errors, while QGVI-I + photo. has three fewer
false accepts. QGVI-I retains the geometric initializer as its default;
photometric refinement remains an optional ablation. These results do not
establish a general benefit or failure of photometric refinement. Coordinate
preservation applies to each acceptance gate, not to the photometric optimizer.

\section{Fixed-Pose Acceptance Audit}
\label{sec:supp_fixed_gate}

Table~\ref{tab:selective_pose} isolates rejection from pose estimation
in QGVI-I, the geometric default.
Every retained camera matrix is unchanged, and every initializer pose
correct at either threshold remains accepted on this cohort. The precision
gain therefore comes from removing incorrect outputs, including
out-of-map returns, rather than improving pose coordinates.

\begin{table}[t]
  \centering
  \caption{\textbf{Selective acceptance at fixed poses.}
  The two arms use the same LoGS geometric estimates as QGVI-I,
  without photometric refinement, and
  are evaluated on 3,678 development queries; the ungated arm includes
  seven retrieval fallbacks. Pose success uses all 1,839 in-map queries;
  precision uses all accepted outputs, including distractors. Pooled
  percentages; the gate changes acceptance only. Bold marks better rates
  and ties.}
  \label{tab:selective_pose}
  \footnotesize
  \setlength{\tabcolsep}{3pt}
  \begin{tabular}{@{}lrr@{}}
    \toprule
    Metric & Ungated & Gated \\
    \midrule
    Accepted outputs ($n$) & 3,678 & 1,831 \\
    False accepts / out-of-map & 1,839/1,839 & 13/1,839 \\
    False accept rate $\downarrow$ & 100.000 & \textbf{0.707} \\
    False rejects / in-map & 0/1,839 & 21/1,839 \\
    False reject rate $\downarrow$ & \textbf{0.000} & 1.142 \\
    \midrule
    Pose success, 1\,m/2$^\circ$ $\uparrow$ & \textbf{96.52} & \textbf{96.52} \\
    Pose success, 5\,m/5$^\circ$ $\uparrow$ & \textbf{96.68} & \textbf{96.68} \\
    Precision, 1\,m/2$^\circ$ $\uparrow$ & 48.26 & \textbf{96.94} \\
    Precision, 5\,m/5$^\circ$ $\uparrow$ & 48.34 & \textbf{97.11} \\
    \bottomrule
  \end{tabular}
\end{table}

\section{Historical Text Retrieval and Verification Controls}
\label{sec:historical_text_controls}

QGVI-T here denotes historical fusion.

\begin{table}[!htbp]
  \centering
  \caption{\textbf{Text component ablations.} (a) Pooled relevance-only Top-5
  bounds (\%) for 176 DeepSeek-authored/judged development queries.
  (b) Equal-scene joint Top-1/Top-5 success for 180 queries: blind
  Qwen3-VL-8B-Instruct relevance and DeepSeek validity reviews;
  source support reviewed separately.
  Bold marks maximal endpoints (a) or success (b);
  overlapping bounds preclude ranking.}
  \label{tab:text_ablation}
  \small
  \textbf{(a) Retrieval and verification}\par\smallskip
  \begin{tabular}{@{}lccc@{}}
    \toprule
    Retrieval & Visual verifier & $n$ & \shortstack{Relevant Top-5\\success (\%) $\uparrow$} \\
    \midrule
    CLIP & No & 176 & 68.2--97.2 \\
    CLIP & Yes & 176 & 76.1--81.8 \\
    Captions (QGVI-T) & No & 176 & 80.7--\textbf{99.4} \\
    Captions & Yes & 176 & \textbf{84.1}--88.1 \\
    \bottomrule
  \end{tabular}
  \par\medskip
  \textbf{(b) Joint relevance and validity on the 180-query cohort}\par\smallskip
  \begin{tabular}{@{}lcc@{}}
    \toprule
    Method & Top-1 (\%) $\uparrow$ & Top-5 (\%) $\uparrow$ \\
    \midrule
    QGVI-T & 58.3 & \textbf{78.3} \\
    QGVI-T + verification & \textbf{66.1} & 72.8 \\
    CityLoC, without refinement & 25.0 & 30.7 \\
    CityLoC, with refinement & 26.1 & 34.6 \\
    \bottomrule
  \end{tabular}
\end{table}

\begin{table}[!htbp]
  \centering
  \caption{\textbf{Scene and query-type breakdown.} Relevance-only Top-5
  bounds (\%) for 176 exposed DeepSeek-judged queries ($n$: unique IDs).
  Author-assigned types lack independent source grounding;
  answerability/validity are unverified. Bold marks each row's maximal endpoints,
  including ties; overlapping bounds preclude ranking.}
  \label{tab:breakdown}
  \small
  \setlength{\tabcolsep}{2pt}
  \footnotesize
  \begin{tabular}{@{}lrcc@{}}
    \toprule
    Scene / query type & $n$ & \shortstack{CLIP + verifier\\Top-5 (\%) $\uparrow$} & \shortstack{QGVI-T + verification\\Top-5 (\%) $\uparrow$} \\
    \midrule
    Block Small & 20 & \textbf{85.0} & \textbf{85.0} \\
    Block A & 20 & 80.0--\textbf{85.0} & \textbf{85.0} \\
    NYC & 20 & \textbf{90.0} & \textbf{90.0} \\
    SF & 20 & 80.0--95.0 & \textbf{85.0}--\textbf{100.0} \\
    Citysample & 20 & 65.0--70.0 & \textbf{80.0} \\
    Colosseum & 16 & 75.0--87.5 & \textbf{87.5}--\textbf{100.0} \\
    Elvenruin & 20 & 60.0--75.0 & \textbf{75.0}--\textbf{85.0} \\
    Park & 20 & 90.0 & \textbf{95.0} \\
    Road & 20 & 60.0 & \textbf{75.0} \\
    \midrule
    Object & 61 & 88.5 & \textbf{91.8}--\textbf{95.1} \\
    Attribute & 35 & 80.0--85.7 & \textbf{91.4} \\
    Spatial relation & 35 & 71.4 & \textbf{85.7} \\
    Counting & 34 & 64.7--\textbf{88.2} & \textbf{73.5}--\textbf{88.2} \\
    Visible text & 11 & \textbf{45.5} & \textbf{45.5} \\
    \midrule
    All queries & 176 & 76.1--81.8 & \textbf{84.1}--\textbf{88.1} \\
    \bottomrule
  \end{tabular}
\end{table}

On 176 queries, verification yields 148--155 Top-5 successes
with captions and 134--144 with CLIP, versus 142--175 for captions alone
(Table~\ref{tab:text_ablation}); overlapping bounds preclude a verifier-effect
claim. Verification gives 45/180 empty returns.
Counting is untested: 30/34 descriptions lack target counts
(Table~\ref{tab:breakdown}).

\section*{Acknowledgments}
ChatGPT (OpenAI) assisted with manuscript drafting and revision, code development, and experimental workflow support. The authors are responsible for the methods, results, and conclusions.

\end{document}